%% file: main.tex
\documentclass[11pt,letterpaper]{article}
\usepackage[margin=1in]{geometry}
\usepackage[T1]{fontenc}
\usepackage[utf8]{inputenc}
\usepackage{times}
\usepackage{microtype}
\usepackage{amsmath,amssymb,mathtools}
\usepackage{graphicx,booktabs,tabularx,array,longtable}
\usepackage{xcolor,colortbl}
\usepackage[font=small,labelfont=bf]{caption}
\usepackage{algorithm,algpseudocode}
\usepackage{listings}
\usepackage[authoryear,round]{natbib}
\setcitestyle{citesep={;},aysep={,},yysep={;}}
\usepackage{url}
\usepackage[hidelinks]{hyperref}
\hypersetup{pdftitle={VisualCoding: Executable Skills for Reliable Agentic Visual Reasoning}}

\definecolor{vcblue}{HTML}{3266A8}
\definecolor{vcteal}{HTML}{14807C}
\definecolor{methodshade}{HTML}{EFEDFA}
\definecolor{gaincolor}{HTML}{008744}
\definecolor{losscolor}{HTML}{C62828}
\definecolor{neutralcolor}{HTML}{666666}
\definecolor{draftcolor}{HTML}{9A3C72}
\newcommand{\method}{\textsc{VisualCoding}}
\newcommand{\ind}{\mathbb{I}}

\title{VisualCoding: Learning Executable Skills for \\
Reliable Visual Reasoning}
\author{
  Jiaqi Liu\textsuperscript{1,2}
  \quad Ruinan Jin\textsuperscript{1,2}
  \quad Xiaoxiao Li\textsuperscript{1,2}
  \\[0.5em]
  \small \textsuperscript{1}Department of Electrical and Computer Engineering, University of British Columbia
  \\
  \small \textsuperscript{2}Vector Institute
  \\[0.3em]
}

\date{}

\begin{document}
\maketitle
\begin{abstract}
Learning reusable skills can improve visual reasoning agents, yet applying a skill still requires the agent to implement or invoke a procedure and produce the final answer. Errors and hallucinations at these steps can prevent acquired experience from translating into reliable single-run performance. 
We propose \method{}, which combines execution-guided acquisition of reusable visual programs with direct adoption of their answers on applicable inputs. \emph{Execution-Guided Program Acquisition} composes visual primitives and additional code into family-level solvers, then iteratively tests and revises them on labeled training instances. \emph{Scope-Aware Direct Execution} runs a retained solver and returns its valid in-scope answer. 
We evaluate \method{} on four models across four benchmarks spanning 69 task families. \method{} exceeds the strongest evaluated baseline in mean accuracy by 4.45 percentage points with Gemini-3.1-Pro and by 14.78 points when reusing the frozen Pro-acquired artifacts with Flash-Lite. Ablations support execution-guided acquisition and show that direct execution outperforms textual or callable reuse of the same programs.
Code is available at \href{https://github.com/jiaq-liu/VisualCoding}{https://github.com/jiaq-liu/VisualCoding}.
\end{abstract}
\suppressfloats[t]

\input{sections/introduction}

\input{sections/related_work}
\input{sections/method}
\input{sections/experiments}
\input{sections/discussion}
\input{sections/conclusion}

\bibliography{references}
\bibliographystyle{plainnat}
\clearpage
\appendix
\input{sections/appendix}
\end{document}

%% file: sections/introduction.tex
\section{Introduction}
\label{sec:intro}

\begin{figure}[tbp]
  \centering
  \includegraphics[width=\linewidth]{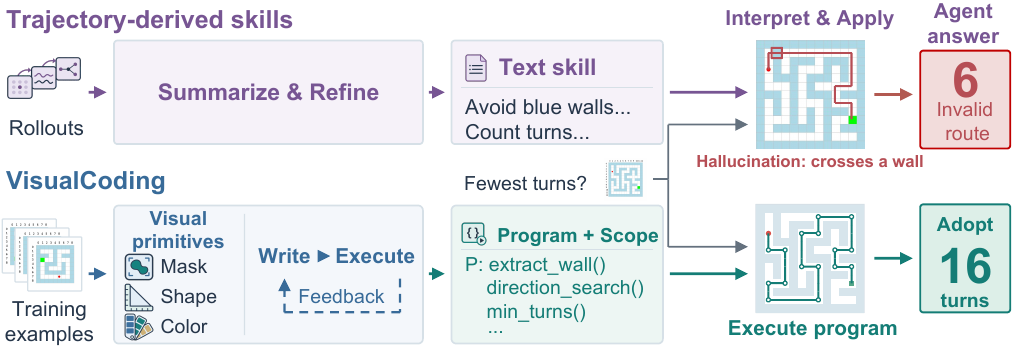}
  \caption{\textbf{From text skills to executable experience.} Agents can still make errors when applying text skills distilled from rollouts (top). \method{} refines reusable programs through execution feedback and directly adopts their valid answers for more reliable visual reasoning (bottom).}
  \label{fig:teaser}
\end{figure}

Agent skills allow reasoning agents to retain and reuse problem-solving experience across tasks. Existing methods distill or refine instructions, code templates, and scripts from execution trajectories without updating model parameters \citep{jiang2026xskill,yang2026skillopt,ni2026trace2skill}. These skills create reusable textual prompts that guide subsequent tool use and reasoning, reducing the need to develop each solution from scratch. However, their effectiveness depends on how agents implement the retained procedures and use their results to answer new questions.

For visual reasoning, however, trajectory-derived textual skills still leave critical execution decisions and final-answer generation to the agent. Even with visual tools and coding interfaces, applying a skill requires translating its guidance into measurements and computations that respect the current image's spatial constraints~\citep{zhao2025pyvision,hong2025deepeyesv2,zhao2026pyvisionrl}. Figure~\ref{fig:teaser} illustrates this gap: the injected skill explicitly requires avoiding blue wall cells, yet the agent describes a wall-crossing route and answers 6. The skill states the correct constraint, but the agent fails to enforce it in the proposed path. Even when a computation yields a correct result, final-answer generation creates another chance for hallucination or misinterpretation to produce an incorrect response. Text skills therefore leave room for errors both in executing a visual strategy and in reporting its result.

These failures are not inevitable. Many visual reasoning tasks can be expressed as compositions of visual primitives, including spatial transformations, image enhancement, and element matching. Within a task family, exploring these compositions can turn abstract textual guidance into executable solution procedures. Labeled training images provide a basis for this exploration: candidate programs can be executed across instances and revised using correctness feedback, execution errors, and intermediate diagnostics. Such iterative exploration can uncover visual computations absent from the initial attempts and preserve them as reusable implementations, extending experience acquisition beyond summarizing existing trajectories.

Building on this insight, we propose \method{}, a framework that acquires deployable solvers for known visual task families and directly adopts their valid in-scope answers (Figure~\ref{fig:teaser}). \textbf{Execution-Guided Program Acquisition} combines visual primitives with additional Python code to construct a shared solver for each task family. It iteratively tests and revises candidate solvers on training instances, checking changes to visual measurements, matching criteria, and search procedures for both gains and regressions. Training-side scope fitting and selection retain an eligible solver and its empirical applicability rule, both of which are frozen before deployment. \textbf{\mbox{Scope-Aware Direct Execution}} runs the frozen solver and directly returns its answer when validity and scope checks pass. This direct path preserves the computed result without requiring the deployment agent to reconstruct the computation, decide whether to invoke the solver, or generate the final answer. Rejected outputs fall back to the base agent.
For task families without a deployable program, we distill text skills from available execution trajectories for use on subsequent test instances.

Our contributions are:
\begin{itemize}
  \item We introduce \method{}, which couples execution-guided acquisition of deployable visual solvers with direct answer adoption on valid in-scope inputs. It refines executable implementations across training instances and reuses the retained computation without requiring the deployment agent to reconstruct it or decide whether to adopt its answer.

 \item Across four benchmarks, \method{} achieves the best mean accuracy among evaluated adaptations with Gemini-3.1-Pro and in transfer to Flash-Lite. Frozen Flash-Lite routes also improve DeepSeek and GPT deployment.

 \item Ablations show higher mean accuracy from \method{} than from textual or callable reuse of the same programs. On some task families, single-run accuracy also exceeds the base agent’s pass@4, suggesting that iterative program refinement can learn solutions missed by repeated sampling
\end{itemize}

%% file: sections/related_work.tex
\section{Related Work}
\label{sec:related}

\subsection{Experience and Skill Learning}

Agent Workflow Memory extracts reusable workflows, Dynamic Cheatsheet accumulates problem-solving strategies, and Voyager retains experience from exploration in Minecraft \citep{wang2024awm,suzgun2025dynamiccheatsheet,wang2023voyager}. ReasoningBank distills successful and failed trajectories into reasoning memory, ACE evolves contextual playbooks, and MetaFlowLLM organizes experience into hierarchical workflows \citep{ouyang2026reasoningbank,zhang2026ace,fan2025metaflow}.

Agent Skills organize procedural knowledge for reuse, including multimodal guidance in XSkill and MMSkills~\citep{jiang2026xskill,zhang2026mmskills}, trajectory distillation in Trace2Skill \citep{ni2026trace2skill}, and skill or context optimization in SkillOpt and Meta Context Engineering \citep{yang2026skillopt,ye2026mce}. SkillGen refines skills through paired evaluations with and without the skill, while SkillsBench evaluates their utility across tasks \citep{ma2026skillgen,li2026skillsbench}. Skill-Pro (ProcMEM) learns procedural memory without updating model weights \citep{mi2026skillpro}. Executable reuse includes polymorphic skills in PolySkill, logic-grounded programs in NSI, and verified tool creation or compilation in RefTool and EvoC2F \citep{yu2026polyskill,shao2026nsi,liu2026reftool,wei2026evoc2f}. Retaining reusable experience or text skills does not ensure a correct final answer: model-mediated implementation, invocation, or answer generation can still introduce execution errors or hallucinations. 
We instead encode the acquired computation in an executable solver and directly adopts its valid in-scope output.
\subsection{Agentic Visual Reasoning}

VisProg and ViperGPT compose visual modules into programs, while PyVision constructs tools dynamically during inference \citep{gupta2022visprog,suris2023vipergpt,zhao2025pyvision}. PIPS introduces program-based computation alongside chain-of-thought reasoning \citep{stein2025pips}. DeepEyes, Pixel Reasoner, PyVision-RL, VLM-R$^3$, and AdaReasoner use reinforcement learning to improve visual tool use and orchestration \citep{zheng2025deepeyes,wang2025pixelreasoner,zhao2026pyvisionrl,jiang2025vlmr3,song2026adareasoner}.
TVP abstracts solved programs into validated tools for subsequent program generation, while Code-with-Image refines textual skills through executable reflection \citep{wu2025tvp,yang2026cwi}. Perception Programs convert tool outputs into structured cues for further model reasoning \citep{janjua2026perceptionprograms}. 
Existing coding-based approaches primarily construct tools or synthesize solutions for individual inputs. VisualCoding refines a task-family solver through execution feedback across labeled training instances and freezes it for reuse. Its valid in-scope outputs directly supply the final answer.

%% file: sections/method.tex
\section{Method}
\label{sec:method}

\begin{figure}[t]
  \centering
  \includegraphics[width=\linewidth]{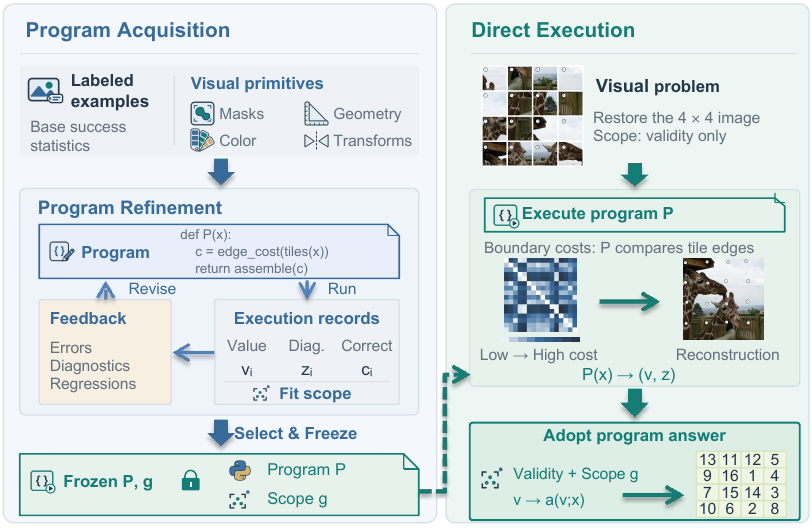}
  \caption{\textbf{\method{} overview.} \textbf{Program acquisition:} training examples, baseline statistics, and visual primitives support program search. Execution feedback guides revision, and scope fitting and selection retain a frozen program $P$ and scope $g$. \textbf{Direct execution:} valid in-scope outputs directly supply the answer through an adapter.
  }
  \label{fig:overview}
\end{figure}

\method{} acquires reusable visual programs from training samples and deploys them on applicable inputs. As shown in Figure~\ref{fig:overview}, in \textit{program acquisition}, a writer composes visual primitives and Python code, then revises candidates using execution feedback. Training-side scope fitting and selection retain a program with an empirical scope rule. In \textit{scoped execution}, the frozen program runs on a new image and question, and its answer is returned when validity and scope checks pass.

\subsection{Overview and Executable Skills}
\label{sec:setup}
\label{sec:primitives}

For a VQA, let $x=(I,q)$ contain an image and the full question text, and let $y$ be the reference answer. For each known task family $f$, acquisition uses labeled examples $\mathcal D_f^{\mathrm{tr}}$ and outcome from a base agent $\pi_0$. We denote its empirical single-run accuracy on training example $i$ by $b_i$, estimated from independent rollouts 
. Labels supervise acquisition and scoring but are excluded from program inputs to prevent label-based shortcuts.

An executable skill consists of a solver $P_f$, an empirical scope rule $g_f$, and an answer adapter $a_f$. Its interface is
\begin{equation}
 P_f(x;\mathcal L)=(v,z),\qquad \tilde y=a_f(v;x),
 \label{eq:contract}
\end{equation}
where $\mathcal L$ is a fixed visual primitive library, $v$ is the computed value, and $z$ contains diagnostics. The adapter converts $v$ to the task's answer convention. The scope rule uses program outputs and label-free task information to decide whether to adopt $\tilde y$. We retain at most one deployable program per family, which may contain internal routes.

The writer combines visual primitives for masks, components, color, geometry, and periodicity with ordinary Python control flow and numerical libraries. Perception interfaces support detection and model queries for discrete perceptual inputs. Acquisition learns the composition, additional code, and scope while keeping the primitives fixed. 

Diagnostics support both revision and applicability checking. For example, periodicity estimation returns a period and its strength. During acquisition, diagnostics, substep checks, and intermediate images help locate failed computations. The scope fitter uses selected diagnostics to identify inputs on which to adopt the program's answer.

\subsection{Execution-Guided Program Acquisition}
\label{sec:acquisition}

\paragraph{Generate, execute, and revise.}
Inspired by counterexample-guided inductive synthesis \citep{solarlezama2008sketching}, we develop a computation shared by a task family through repeated execution and revision. It starts from a profile of labeled training examples, baseline statistics, and bounded image probes, then proposes a program, an answer convention, and diagnostic features. The harness executes each candidate and returns per-example correctness, execution errors, diagnostics, and intermediate visualizations. Subsequent proposals receive the current best candidate, this feedback, and a ledger of previous attempts. Comparisons with the best candidate identify newly solved examples and regressions, guiding repairs and extensions (Algorithm~\ref{alg:acquire}). All feedback only comes from training instances.

\paragraph{Fit an empirical scope rule.}
For a fixed candidate, let $c_i=\ind[\tilde y_i=y_i]$ denote correctness under the task evaluator and let $s_i(g)$ indicate a valid answer accepted by rule $g$. Within each program-declared route, or the full family without routing, we search a compact rule class $\mathcal G$ for maximum training coverage subject to minimum precision $\tau$ and support $n_{\min}$:
\begin{equation}
 \max_{g\in\mathcal G}\ \sum_i s_i(g)
 \quad\text{s.t.}\quad
 \frac{\sum_i s_i(g)c_i}{\sum_i s_i(g)}\geq\tau,
 \qquad \sum_i s_i(g)\geq n_{\min}.
 \label{eq:scopefit}
\end{equation}
Rules combine numeric, Boolean, or categorical diagnostic comparisons. Required features that are missing, unsupported routes, invalid answers, and execution errors trigger fallback. Appendix~\ref{app:scope} specifies the rule class and threshold search.

After synthesizing the program, we check whether its scope rule depends too heavily on individual training tasks. We evaluate the scope rule by leaving out one task at a time, refitting the rule on the stored program outputs from the remaining tasks, and measuring precision and coverage on the left-out task. To assess rule stability, we measure how often the refitted rule makes the same scope decisions as the rule fitted on all training outputs. The program remains fixed throughout these checks.

\paragraph{Select and freeze.}
The high-precision tier requires training and leave-one-out precision, sufficient support, and rule stability. Otherwise, a gain tier applies its own precision, support, and stability checks and requires a positive bootstrap lower bound on paired improvement over the base-agent reference $b_i$. Within a run, high-precision candidates rank ahead of gain-tier candidates. Deployment additionally requires at least 50\% training coverage. We freeze the best eligible program and scope under the training-side selection rule. Acquisition budgets, stopping criteria, and ranking details appear in Appendix~\ref{app:scope} and~\ref{app:selection}. If no program is eligible, a separate summarizer builds a text skill from baseline trajectories and labeled outcomes.
Skill construction and formatting checks are specified in Appendix~\ref{app:textskill}.

\begin{algorithm}[htbp]
\caption{Execution-guided acquisition within one family run}
\label{alg:acquire}
\begin{algorithmic}[1]
\Require Labeled training set $\mathcal D^{\mathrm{tr}}$, base statistics $b$, writer $W$, library $\mathcal L$
\State $F\gets\mathrm{ProfileAndProbe}(\mathcal D^{\mathrm{tr}})$
\State $\mathcal I^\star\gets\varnothing$ \Comment{Current best candidate}
\State feedback $E\gets\varnothing$; ledger $H\gets\varnothing$
\While{the search stopping rule is not met}
  \State $(P,a,Z)\gets W(F,b,\mathcal L,\mathcal I^\star,E,H)$ \Comment{Propose or revise}
  \State $R\gets\mathrm{ExecuteAndScore}(P,a,\mathcal D^{\mathrm{tr}},\mathcal L)$
  \State $g\gets\mathrm{FitScope}(R,Z,b)$ \Comment{Use declared diagnostics $Z$}
  \State $m\gets\mathrm{EvaluateCandidate}(P,g,a,R,b)$ \Comment{Evaluation metadata}
  \State $\mathcal I\gets(P,g,a,m)$
  \State $E\gets\mathrm{Feedback}(R,m,\mathcal I^\star)$ \Comment{Errors, diagnostics, regressions}
  \State $H\gets\mathrm{AppendAttempt}(H,\mathcal I,E)$
  \State $\mathcal I^\star\gets\mathrm{KeepBest}(\mathcal I^\star,\mathcal I)$
\EndWhile
\State \Return $\mathcal I^\star$ \Comment{Deployment eligibility is checked across runs}
\end{algorithmic}
\end{algorithm}

\subsection{Scope-Aware Direct Execution}
\label{sec:deployment}

Deployment uses selective prediction with agent fallback \citep{geifman2017selective}. Each family has a frozen route $r_f$: a deployable program takes priority, followed by a text skill $S_f$ when available, and otherwise the base agent. On the program route, we execute $P_f$ and let $s_f(x)$ indicate that its answer is valid. For the collection of frozen family routes $\mathcal A$, the policy is
\begin{equation}
 \pi_{\mathcal A}(x)=
 \begin{cases}
   a_f(v;x), & r_f=\mathrm{program}\text{ and }s_f(x)=1,\\
   \pi_0(x\mid S_f), & r_f=\mathrm{skill},\\
   \pi_0(x), & \text{otherwise}.
 \end{cases}
 \label{eq:policy}
\end{equation}
An accepted program answer is returned directly. A rejected output falls back to the base agent without the program output or a text skill. Text skills enter the agent's system prompt only for families without deployable programs. Routes and scope rules remain frozen during evaluation. 
Appendix~\ref{app:policy_gain} separates the policy's program and text-skill contributions.

%% file: sections/experiments.tex
\section{Experiments}
\label{sec:experiments}

\subsection{Experimental Setup}
\label{sec:exp_setup}

\paragraph{Benchmarks and supervision.}
We evaluate on four widely used benchmarks introduced at top-tier conferences: TIR-Bench, MIRA, GameQA, and PuzzleVQA \citep{li2025tirbench,zhou2025mira,tong2025game,chia2024puzzlevqa}. Our evaluation covers 69 task families and 2,960 test questions.
Appendix~\ref{app:development} specifies the training/test partitions.

\paragraph{Models and baselines.}
Gemini-3.1-Pro authors the retained artifacts, which we evaluate with Gemini 3.1 Pro, Gemini 3.1 Flash-Lite, DeepSeek V4.1 Flash, and GPT-5.4-mini. Gemini-3.1-Pro and Flash-Lite use executor-specific training-side deployment checks. DeepSeek and GPT reuse frozen Flash-Lite (FL) routes without target-model training or recalibration. Program readings are shared across executors; Appendix~\ref{app:main_accounting} specifies the evaluation protocol.

We compare XSkill, SkillOpt, Trace2Skill, GEPA, and TextGrad \citep{jiang2026xskill,yang2026skillopt,ni2026trace2skill,agrawal2025gepa,yuksekgonul2024textgrad}. XSkill, SkillOpt, Trace2Skill, and \method{} share the same tool suite. GEPA and TextGrad use tool-free inference. 

\paragraph{Evaluation.}
Main comparisons and ablations use the same rollout for each base-agent or text-skill fallback and frozen program readings; each agent intervention uses one rollout. Dataset accuracy weights tasks, and Avg. equally weights the four datasets. The sampling analysis separately uses four recorded base-agent attempts per question. Agent answers are scored by model judges. Complete family-level scores appear in Appendix~\ref{app:family_results}.
\label{sec:cost}

\subsection{Main Comparison}
\label{sec:main_results}

With Gemini-3.1-Pro, \method{} improves single-run accuracy over the tool agent on all four benchmarks (Table~\ref{tab:main}). \method{} raises mean accuracy from 66.78\% to 76.11\% ($+9.33$ points), compared with 71.67\% for XSkill, the strongest external average under the evaluated adaptations. The largest gains are on TIR-Bench and MIRA, where the base agent leaves more questions unsolved, while PuzzleVQA rises from an already high 93.72\% to 98.58\%. The ablations below examine which parts of acquisition and deployment contribute to these gains.

\input{sections/main_results_table}

\subsection{Cross-Executor Reuse}
\label{sec:transfer}
In addition, we compare the effectiveness of different methods for transferring the training outputs of Gemini-3.1-Pro to the less powerful Gemini-3.1-Flash-Lite. \method{} raises mean accuracy from the tool agent's 43.42\% to 68.57\% ($+25.14$ points), compared with 53.78\% for XSkill (Table~\ref{tab:transfer}(a)). The program branch preserves its stored computation when the deployment agent changes, while text and base-agent branches continue to depend on the target executor. This setting uses Flash-Lite-calibrated routes, so the larger gain than with Pro combines executor differences with route calibration. Appendix~\ref{app:main_accounting} compares this configuration with the frozen Pro deployment policy to examine reuse without Flash-Lite-specific deployment calibration.

Panel (b) tests reuse of the frozen FL routes with DeepSeek and GPT, without training rollouts or deployment checks for either target. DeepSeek improves from 60.74\% to 71.66\%, and GPT from 34.70\% to 63.70\% ($+10.92$ and $+29.00$ points). Both targets use the same configuration across all four datasets. These gains demonstrate reuse of the retained artifacts and stored readings beyond the source executor, under each target's scoring protocol.

\input{sections/transfer_results_table}

\subsection{Success Beyond the Sampling Pool}
\label{sec:observed_passk}

\method{} explores and refines reusable programs to improve single-run reliability, which enables performance that even surpasses that of multiple independent attempts. On the 746-question TIR-Bench panel, its estimated pass@1 exceeds the base agent's observed pass@4 in four of twelve task families (Table~\ref{tab:tir_sampling_families}), including maze (100\% vs.\ 95\%), OCR (87.5\% vs.\ 72.5\%), and rotation game (100\% vs.\ 98\%).

Of the 194 questions on which all four base attempts fail, acquired programs take over 88 and answer 81 correctly. On the 211 intermittently solved questions, program-only policy accuracy reaches 62.80\%, compared with the base agent's 52.25\% estimated pass@1. These results suggest that deeper exploration and refinement within task families can improve single-run reliability enough to outperform repeated independent attempts in some families.


\subsection{Execution Cases and Supporting Analyses}
\label{sec:case_study}

\begin{figure}[!t]
\centering
\includegraphics[width=\linewidth]{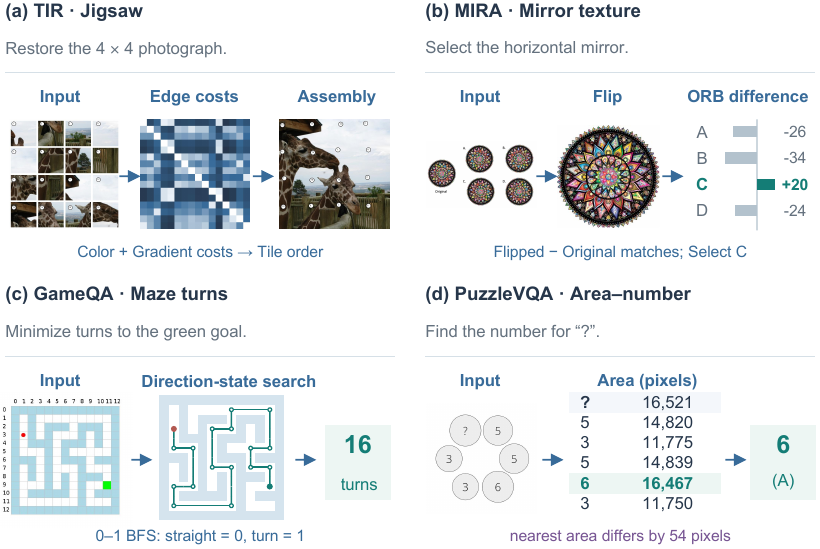}
\caption{\textbf{Representative cases.}
(a) TIR-Bench: lighter cells indicate lower boundary-matching costs.
(b) MIRA: scores show the difference between reflected and original ORB match counts; C has the largest increase.
(c) GameQA: the displayed path has 16 turns and is verified on the extracted grid.
(d) PuzzleVQA: measured component areas identify six (A) as the answer.}
\label{fig:cases}
\end{figure}

Figure~\ref{fig:cases} presents successful representative cases on four benchmarks. On TIR-Bench, the program extracts puzzle pieces, generates candidate assemblies using boundary compatibility, and ranks complete layouts to recover the image. On MIRA, it compares feature matches before and after reflection to identify the mirrored pattern. On GameQA, it converts the maze into a traversable grid and searches position--direction states to minimize turns. On PuzzleVQA, it matches measured circle areas to their associated numbers, using one model call to transcribe characters. Together, these examples illustrate how visual measurements become answers through task-specific scoring, search, and matching procedures.

The baseline traces illustrate why trajectory-based text skill fail in execution. XSkill and Trace2Skill compute local boundary matches but then manually specify puzzle layouts without optimizing the complete arrangement; SkillOpt repeatedly visualizes proposed layouts without executing its suggested matching procedure. In the other cases, unsuccessful runs misidentify petal order, describe a maze route that crosses an obstacle, or substitute an assumed numerical relation for size measurement. These failures show that relevant high-level guidance can leave critical implementation details and visual judgments to the deployment agent and lead to failure. Our acquisition instead repeatedly tests and refines a complete procedure across a task family. Execution feedback guides revisions to preprocessing, matching criteria, candidate generation, and search, while comparisons across training instances identify improvements and regressions. The retained program therefore captures the implementation details needed to connect visual measurements to the answer. Reusing this procedure reduces reliance on per-input implementation and repeated visual interpretation, while direct execution preserves the computed result instead of subjecting it to another model judgment. Together, these choices reduce opportunities for visual hallucination during deployment. The same-program ablations in Section~\ref{sec:ablations} examine the benefits of retaining and directly executing these computations. 

\subsection{Ablation Studies}
\label{sec:ablations}

\begin{table}[tbp]
\centering
\small
\setlength{\tabcolsep}{8pt}
\caption{\small\textbf{Acquisition and deployment ablations with Gemini-3.1-Pro.} Complete-panel accuracy (\%): TIR-Bench has 746 questions and MIRA 363. Panel (a) compares feedback conditions on five re-acquired families under controlled deployment settings. Panel (b) changes how frozen main-experiment programs are used.  Both panels retain full-system outputs on unaffected questions. Feedback-granularity and deployment-configuration controls appear in the appendix.}
\label{tab:ablations}
\begin{tabular}{lcc}
\toprule
Variant & TIR-Bench & MIRA\\
\midrule
Tool agent & 62.9 & 37.2\\
\midrule
\multicolumn{3}{l}{\emph{(a) Acquisition}}\\
Full execution feedback & 75.9 & 46.6\\
Independent search & 73.5 & 46.0\\
No scope during acquisition & 68.3 & 42.0 \\
\midrule
\multicolumn{3}{l}{\emph{(b) Deployment}}\\
\multicolumn{3}{l}{\emph{Frozen main-experiment programs}}\\
Program-derived text procedure & 69.4 & 44.4\\
Callable program & 73.2 & 44.9\\
Reading injection & 75.5 & 45.5\\
\rowcolor{methodshade}Full method & 76.3 & 47.1\\
\bottomrule
\end{tabular}
\end{table}


\paragraph{Acquiring a reusable computation.}
Table~\ref{tab:ablations}(a) compares execution-guided revision with independent candidate search on TIR-Bench and MIRA using Gemini-3.1-Pro. Since at least five families have available programs across the benchmark, the feedback comparison re-acquires five families per benchmark, sharing initial facts, candidate slots, and selection rules. On these five families, text-skill fallback and the additional deployment check are disabled, and failed acquisitions revert to the tool agent; all other families retain their full-system outputs. Full feedback supplies earlier candidates, per-example outcomes, and diagnostics; independent search omits feedback from previous candidates. The no-scope acquisition arm retains full execution feedback while omitting scope fitting during program acquisition. All three conditions report mean accuracy over three independent acquisition runs. Compared with independent search, full feedback yields higher complete-panel accuracy: 75.9\% versus 73.5\% on TIR and 46.6\% versus 46.0\% on MIRA. The full-feedback reference therefore uses different artifacts and deployment settings from the full method in panel (b). The no-scope arm achieves a mean complete-panel accuracy of 68.3\% on TIR-Bench and 42.0\% on MIRA.

\paragraph{Reusing the acquired implementation.}
Panel (b) holds the main experiment's acquired programs and deployed routes fixed while changing how the agent receives the computation. Direct execution in the full method reaches 76.3\% on TIR and 47.1\% on MIRA. Supplying the acquired programs as textual procedures yields 69.4\% and 44.4\%, respectively, while exposing them as callable code yields 73.2\% and 44.9\%. Callable code preserves the implementation, but leaves invocation and answer adoption to the agent. 
These controls compare deployment forms of the same acquired experience, showing that converting the program into text or enabling the agent to perform selective call introduces additional errors.

\paragraph{Preserving computed answers.}
Reading injection supplies the same frozen reading for a tool-free agent continuation, isolating answer use from program construction and invocation. Accuracy falls to 75.5\% on TIR and 45.5\% on MIRA. Among the 268 affected TIR takeovers, continuation changes 10 correct readings into incorrect answers and repairs four incorrect readings. On MIRA's 71 affected takeovers, the corresponding counts are seven and one. In each panel, six net lost answers account for the observed full-panel difference. These transitions identify answer continuation as a source of deployment loss after the computation is already available.

%% file: sections/main_results_table.tex
\begin{table}[t]
\caption{\textbf{Main comparison with Gemini 3.1 Pro.} Accuracy (\%); Avg. equally weights four task-weighted dataset accuracies. GQA: GameQA-17; PVQA: PuzzleVQA; TF: tool-free inference. Bold/underline mark best/second-best distinct displayed accuracies within each executor (ties included). Small green/red numbers are gains/losses in percentage points: GEPA/TextGrad versus Direct; XSkill/Trace2Skill/SkillOpt/VC versus Tool agent. Protocol details: Appendix~\ref{app:main_accounting}.}
\label{tab:main}
\centering\setlength{\tabcolsep}{11pt}
\begin{tabular}{lrrrrr}
\toprule
Method & TIR & MIRA & GQA & PVQA & Avg.\\
\midrule
Direct (TF) & 48.9 & 33.3 & 68.3 & 93.6 & 61.0\\
GEPA (TF) & 52.9\textsubscript{\textcolor{gaincolor}{+4.0}} & 35.0\textsubscript{\textcolor{gaincolor}{+1.7}} & 73.7\textsubscript{\textcolor{gaincolor}{+5.4}} & \underline{97.8}\textsubscript{\textcolor{gaincolor}{+4.3}} & 64.9\textsubscript{\textcolor{gaincolor}{+3.8}}\\
TextGrad (TF) & 54.0\textsubscript{\textcolor{gaincolor}{+5.1}} & 33.3\textsubscript{\textcolor{neutralcolor}{0.0}} & 75.1\textsubscript{\textcolor{gaincolor}{+6.8}} & 96.7\textsubscript{\textcolor{gaincolor}{+3.1}} & 64.8\textsubscript{\textcolor{gaincolor}{+3.8}}\\
\midrule
Tool agent & 62.9 & 37.2 & 73.3 & 93.7 & 66.8\\
XSkill & \underline{68.9}\textsubscript{\textcolor{gaincolor}{+6.0}} & 42.4\textsubscript{\textcolor{gaincolor}{+5.2}} & 78.4\textsubscript{\textcolor{gaincolor}{+5.1}} & 96.9\textsubscript{\textcolor{gaincolor}{+3.2}} & \underline{71.7}\textsubscript{\textcolor{gaincolor}{+4.9}}\\
Trace2Skill & 65.8\textsubscript{\textcolor{gaincolor}{+2.9}} & \underline{45.2}\textsubscript{\textcolor{gaincolor}{+8.0}} & 74.9\textsubscript{\textcolor{gaincolor}{+1.6}} & 95.7\textsubscript{\textcolor{gaincolor}{+2.0}} & 70.4\textsubscript{\textcolor{gaincolor}{+3.6}}\\
SkillOpt & 67.6\textsubscript{\textcolor{gaincolor}{+4.7}} & 37.7\textsubscript{\textcolor{gaincolor}{+0.6}} & \underline{78.8}\textsubscript{\textcolor{gaincolor}{+5.4}} & 97.6\textsubscript{\textcolor{gaincolor}{+3.9}} & 70.4\textsubscript{\textcolor{gaincolor}{+3.6}}\\
\rowcolor{methodshade}\method{} & \textbf{76.3}\textsubscript{\textcolor{gaincolor}{+13.4}} & \textbf{47.1}\textsubscript{\textcolor{gaincolor}{+9.9}} & \textbf{82.5}\textsubscript{\textcolor{gaincolor}{+9.1}} & \textbf{98.6}\textsubscript{\textcolor{gaincolor}{+4.9}} & \textbf{76.1}\textsubscript{\textcolor{gaincolor}{+9.3}}\\
\bottomrule
\end{tabular}
\end{table}

%% file: sections/transfer_results_table.tex
\begin{table}[t]
\caption{\textbf{Cross-executor reuse.} Accuracy (\%); annotations follow Table~\ref{tab:main}, ranked separately per executor. (a) External Pro-learned artifacts deploy with Flash-Lite; ours uses Flash-Lite training-side calibration. (b) DeepSeek/GPT reuse frozen FL routes and program readings without target calibration. Judge assignments and evaluation protocol: Appendix~\ref{app:main_accounting}.}
\label{tab:transfer}
\centering\setlength{\tabcolsep}{11pt}
\begin{tabular}{lrrrrr}
\toprule
Method & TIR-Bench & MIRA & GameQA & PuzzleVQA & Avg.\\
\midrule
\multicolumn{6}{l}{\emph{(a) Gemini 3.1 Flash-Lite}}\\
Direct (TF) & 28.3 & 17.9 & 28.6 & 78.8 & 38.4\\
GEPA (TF) & 36.3\textsubscript{\textcolor{gaincolor}{+8.0}} & 18.5\textsubscript{\textcolor{gaincolor}{+0.6}} & 40.9\textsubscript{\textcolor{gaincolor}{+12.3}} & 91.8\textsubscript{\textcolor{gaincolor}{+12.9}} & 46.9\textsubscript{\textcolor{gaincolor}{+8.4}}\\
TextGrad (TF) & 33.5\textsubscript{\textcolor{gaincolor}{+5.2}} & 16.3\textsubscript{\textcolor{losscolor}{-1.7}} & 32.5\textsubscript{\textcolor{gaincolor}{+3.9}} & 84.4\textsubscript{\textcolor{gaincolor}{+5.5}} & 41.7\textsubscript{\textcolor{gaincolor}{+3.3}}\\
\midrule
Tool agent & 39.0 & 20.9 & 30.4 & 83.4 & 43.4\\
XSkill & \underline{48.7}\textsubscript{\textcolor{gaincolor}{+9.7}} & \underline{28.4}\textsubscript{\textcolor{gaincolor}{+7.4}} & 45.1\textsubscript{\textcolor{gaincolor}{+14.8}} & \underline{93.0}\textsubscript{\textcolor{gaincolor}{+9.6}} & \underline{53.8}\textsubscript{\textcolor{gaincolor}{+10.4}}\\
Trace2Skill & 44.8\textsubscript{\textcolor{gaincolor}{+5.8}} & 24.5\textsubscript{\textcolor{gaincolor}{+3.6}} & 40.9\textsubscript{\textcolor{gaincolor}{+10.5}} & 86.9\textsubscript{\textcolor{gaincolor}{+3.5}} & 49.3\textsubscript{\textcolor{gaincolor}{+5.8}}\\
SkillOpt & 48.5\textsubscript{\textcolor{gaincolor}{+9.5}} & 22.0\textsubscript{\textcolor{gaincolor}{+1.1}} & \underline{49.8}\textsubscript{\textcolor{gaincolor}{+19.5}} & \underline{93.0}\textsubscript{\textcolor{gaincolor}{+9.6}} & 53.3\textsubscript{\textcolor{gaincolor}{+9.9}}\\
\rowcolor{methodshade}\method{} & \textbf{68.5}\textsubscript{\textcolor{gaincolor}{+29.5}} & \textbf{41.6}\textsubscript{\textcolor{gaincolor}{+20.7}} & \textbf{65.4}\textsubscript{\textcolor{gaincolor}{+35.0}} & \textbf{98.8}\textsubscript{\textcolor{gaincolor}{+15.4}} & \textbf{68.6}\textsubscript{\textcolor{gaincolor}{+25.1}}\\
\midrule
\multicolumn{6}{l}{\emph{(b) Frozen FL routes; no target-model calibration}}\\
\multicolumn{6}{l}{DeepSeek V4.1 Flash}\\
Tool agent & \underline{59.1} & \underline{33.3} & \underline{60.7} & \underline{89.8} & \underline{60.7}\\
\rowcolor{methodshade}\method{} & \textbf{75.6}\textsubscript{\textcolor{gaincolor}{+16.5}} & \textbf{44.4}\textsubscript{\textcolor{gaincolor}{+11.0}} & \textbf{68.9}\textsubscript{\textcolor{gaincolor}{+8.2}} & \textbf{97.8}\textsubscript{\textcolor{gaincolor}{+8.0}} & \textbf{71.7}\textsubscript{\textcolor{gaincolor}{+10.9}}\\
\multicolumn{6}{l}{GPT-5.4-mini}\\
Tool agent & \underline{33.6} & \underline{20.1} & \underline{31.1} & \underline{53.9} & \underline{34.7}\\
\rowcolor{methodshade}\method{} & \textbf{64.9}\textsubscript{\textcolor{gaincolor}{+31.2}} & \textbf{39.4}\textsubscript{\textcolor{gaincolor}{+19.3}} & \textbf{59.7}\textsubscript{\textcolor{gaincolor}{+28.6}} & \textbf{90.8}\textsubscript{\textcolor{gaincolor}{+36.9}} & \textbf{63.7}\textsubscript{\textcolor{gaincolor}{+29.0}}\\
\bottomrule
\end{tabular}
\end{table}

%% file: sections/discussion.tex
\section{Discussion and Limitation}
\label{sec:discussion}

\paragraph{Acquiring and preserving experience.}
The ablations test acquisition and deployment separately (Table~\ref{tab:ablations}). Feedback-guided revision yields higher complete-panel accuracy than independent search, while direct execution outperforms textual and callable reuse on TIR-Bench and MIRA. Reading injection identifies losses from answer continuation; the broader reuse comparisons also involve implementation and invocation decisions. Programs additionally solve questions missed by all four observed baseline attempts, supporting recovery beyond a finite sampling pool.

\paragraph{Applicability and relative utility.}
Scope governs answer adoption within a family, while the deployment check determines which families use programs. With programs and scopes fixed, Flash-Lite benefits from both Pro and Flash-Lite policies, with higher accuracy under the latter (Appendix~\ref{app:main_accounting}). Larger transfer gains thus reflect both baseline headroom and deployment-policy choices.

\paragraph{Limitations.} Due to API cost constraints, we use Gemini 3.1 Pro as the sole model for the program acquisition pipeline and do not explore other SOTA models in this role. For the same reason, our evidence of larger transfer gains for weaker models is limited to Gemini 3.1 Flash-Lite. Our main DeepSeek and GPT evaluations use the deployment policy calibrated for Flash-Lite. The Pro-calibrated deployment check may be overly conservative when reused with a weaker executor; relaxing it may improve accuracy while admitting programs whose relative benefit is less certain (Appendix~\ref{app:main_accounting}). 

%% file: sections/conclusion.tex
\section{Conclusion}
\label{sec:conclusion}

We present \method{}, a framework for learning and deploying executable skills for visual reasoning. It iteratively refines family-level programs using execution feedback on labeled training instances, then directly adopts their valid in-scope answers. Across four benchmarks, \method{} improves mean single-run accuracy over the tool agent by 9.33 percentage points with Gemini-3.1-Pro and 25.14 points with Gemini-3.1-Flash-Lite. The retained artifacts and stored program readings also improve DeepSeek and GPT deployment. Ablations on TIR-Bench and MIRA support feedback-guided acquisition and show higher mean accuracy from direct execution than from textual or callable reuse of the same programs. Acquired programs also solve questions missed by all four recorded base-agent attempts on some tasks. Together, these findings suggest that learning reusable computations can improve single-run reliability by reducing opportunities for execution errors and hallucinations during deployment.

%% file: sections/appendix.tex
\section{Implementation and Acquisition Details}
\label{app:implementation}

\subsection{Interfaces and Primitive Inventory}

A deployable program is stored as $\mathcal I_f=(P_f,g_f,a_f,m_f)$, where $m_f$ records dependencies and acquisition decisions. The frozen family record $\mathcal A_f=(r_f,\mathcal I_f,S_f)$ identifies its deployment route, program artifact, and text skill when available.

The acquired program exposes \texttt{measure(img, task, lib)} and returns a dictionary with \texttt{value} and \texttt{diag}. The task view contains the full question text, including answer choices when present, and permitted metadata, excluding the reference answer. A separate evaluator applies the answer adapter and compares with the label during acquisition or offline scoring. The adapter can interpret a numeric reading, a choice key, a coordinate, or a sequence according to the frozen answer convention. It is kept identical across deployment ablations. A failed execution or unparseable answer cannot become a direct answer.

The generic library contains 30 documented entries. Table~\ref{tab:primitives} groups the generic entries by purpose. Numerical libraries, including NumPy, OpenCV, PIL, SciPy, and scikit-image, remain available to the program within the runner's restrictions.

\begin{table}[ht]
\caption{The fixed generic primitive library. Diagnostic returns can be included in the program's \texttt{diag}; debug images and substep checks support offline feedback.}
\label{tab:primitives}
\begin{center}
\begin{tabularx}{\linewidth}{@{}>{\raggedright\arraybackslash}p{0.22\linewidth}>{\raggedright\arraybackslash}X@{}}
\toprule
Role & Documented entries\\
\midrule
Image preparation & \texttt{arr}, \texttt{gray}, \texttt{background}, \texttt{fg\_mask}, \texttt{ink\_mask}, \texttt{hue\_mask}, \texttt{stretch}, \texttt{upscale}, \texttt{rotate}\\
Structure and geometry & \texttt{components}, \texttt{fill\_holes}, \texttt{holes}, \texttt{contour\_stats}, \texttt{classify}, \texttt{min\_area\_rect}\\
Color measurements & \texttt{color\_name}, \texttt{n\_colors}, \texttt{color\_share}, \texttt{mean\_color}\\
Spatial patterns & \texttt{projection}, \texttt{period}, \texttt{panels8}, \texttt{apply\_variant}\\
Perceptual interfaces & \texttt{detect}, \texttt{count\_objects}, \texttt{ask}\\
Question processing & \texttt{question}, \texttt{numbers\_in}\\
Execution inspection & \texttt{debug}, \texttt{check}\\
\bottomrule
\end{tabularx}
\end{center}
\end{table}

Programs can use pure code, detectors, or model-backed perception. The representation ablations give text and code controls the same coding access and primitives.

\texttt{ask} makes one model request per invocation.

\subsection{Scope Fitting and Acceptance Criteria}
\label{app:scope}

Base-agent outcome statistics provide the acquisition reference. With $K_{\mathrm{tr}}$ rollouts for training example $i$, its empirical single-run accuracy is
\begin{equation}
 b_i=\frac{1}{K_{\mathrm{tr}}}\sum_{j=1}^{K_{\mathrm{tr}}}\ind[\hat y^{(0)}_{ij}=y_i].
 \label{eq:baseline}
\end{equation}
Equality denotes correctness under the fixed evaluator, with per-example denominators when rollout counts differ. The program writer receives labeled examples, these outcome statistics, and feedback from its candidate executions. Full stored baseline trajectories $\mathcal T_f^{\mathrm{tr}}$, including reasoning text and tool sequences, are inputs only to the separate text-skill summarizer.

For each numeric feature, the fitter sorts unique observed values and enumerates adjacent midpoints, subsampling to at most 30 thresholds per feature. It considers both upper and lower inequalities. If neighboring values are $u_j,u_{j+1}$, the associated separation is
\begin{equation}
 t_j=\frac{u_j+u_{j+1}}{2},\qquad
 m_j=\frac{u_{j+1}-u_j}{\max(u)-\min(u)}.
\end{equation}
Constant numeric features do not generate thresholds. Boolean equality tests and categorical equality/inequality tests are also considered, with categorical enumeration limited to features having at most ten observed values. Features are derived without ground-truth access. Raw output values and answer-option features are excluded by default unless explicitly permitted by the diagnostic declaration.

Each supported route receives a conjunction of up to two learned atoms in the default configuration, in addition to fixed requirements. The empty conjunction is allowed: a valid answer can be accepted without further learned comparisons. Fixed requirements are restricted to Boolean or string equality/inequality predicates so a writer-specified numeric threshold cannot bypass the fitter through that field. This restriction does not prevent program-internal numerical branching. Routes with no feasible rule are omitted from the accepted scope.

The default fitting objective maximizes coverage subject to the tier's precision and support requirements. Ties prefer larger paired gain, fewer learned atoms, higher precision, larger separation, and then a deterministic enumeration order. A gain-maximizing fitting objective is supported as a separate configuration; it is not the default. Rule stability is the fraction of leave-one-task-out folds whose fitted rule exactly matches the rule fitted on all stored training readings.

\begin{table}[ht]
\caption{Acceptance hyperparameters.}
\label{tab:gate_defaults}
\begin{center}
\begin{tabular}{@{}lcc@{}}
\toprule
Criterion & High-precision tier & Gain tier\\
\midrule
Minimum fitted precision & 0.95 & 0.50\\
Minimum fitted support per route & 3 & 5\\
Minimum pooled leave-one-out precision & 0.90 & 0.50\\
Minimum rule stability & 0.80 & 0.50\\
Maximum learned atoms per route & 2 & 2\\
Minimum numeric separation & 0.02 & 0.02\\
Minimum leave-one-out support & Nonempty & 5\\
Positive paired-gain lower bound & Not required & Required\\
\bottomrule
\end{tabular}
\end{center}
\end{table}

A preliminary hypothesis check tests whether the candidate's readings agree with training labels before fitting a selective scope. The standard path requires at least three parsed examples and agreement of at least 0.60, assessed over the pooled readings or a supported route. The gain path permits a five-example, 0.50-agreement criterion. Failed routes can be excluded from subsequent fitting.

For the gain tier, the harness forms paired differences $d_i=c_i-b_i$ on the leave-one-out accepted readings with a base-agent reference. It requires at least five such pairs and a positive one-sided 95\% bootstrap lower bound on their mean, using 2,000 resamples and the implementation's fixed bootstrap seed of zero. This is a training-side selection heuristic under adaptive candidate search. It is not a multiple-search-adjusted guarantee of positive test gain. The same caution applies to the high-precision tier, since synthesis and diagnostic design have already used the training examples.

\subsection{Computational Thresholds and Deployment Eligibility}

Program-internal thresholds define a computation and may be proposed by the writer or computed adaptively from an input. Scope thresholds are fitted by the harness from training readings. Acceptance thresholds specify the candidate gate. A separate family-level coverage floor determines eligibility for deployment under the current adaptive policy. The gate and coverage floor are protocol hyperparameters, distinct from the fitted diagnostic thresholds.

\subsection{Search Budgets and Artifact Selection}
\label{app:selection}

The budgets and selection rules below complement Algorithm~\ref{alg:acquire} by specifying when search stops and which artifact is deployed.

The current simple profile has six base rounds with three candidate slots per round; the solver profile has eighteen base rounds with one candidate each and a higher reasoning setting. The default workflow can stop or extend according to progress and acceptance, so neither configuration implies exactly eighteen evaluated candidates. Production runs can extend after acceptance to improve coverage, while controlled acquisition ablations fix the candidate budget. All probes, repair attempts, invalid proposals, and extensions contribute to acquisition cost. Candidate execution is subject to per-task resource limits.

Within a run, the ranking distinguishes high-precision, gain-tier, and unaccepted candidates, then uses leave-one-out gain, the hypothesis check, parsed accuracy, fitted gain, and coverage. The current selector considers artifacts accepted by the gate whose fitted scope also covers at least 50\% of training examples. It orders eligible artifacts by leave-one-out gain, coverage, and run index. The retained artifact contains code, scope, answer convention, provenance, and dependency configuration. Saved candidates that fail either acceptance or deployment eligibility are not deployed.

The adaptive workflow stops when the latest run yields an eligible program or after three runs, then selects among eligible artifacts. If none is eligible, it uses text-skill fallback when available. The coverage floor controls both stopping and deployment eligibility without modifying the internal acceptance tiers.

\subsection{Trajectory-Derived Skills and Frozen Family Routes}
\label{app:textskill}

The text-skill summarizer filters stored rollouts by source, family, executor model, and training task IDs. The main protocol uses Gemini-3.1-Pro experience and a Gemini-3.1-Pro summarizer. It reads saved \texttt{trajectory\_text}, removing the system prompt and adding outcome, final answer, reference answer, and termination status. It does not replay all intermediate image files. Per-task analysts inspect summaries, the question, label, and original image. Successes and failures yield up to three entries per applicable analyst. Tips are fused in batches of twelve and refined once, using temperature $0.3$, top-$p=1$, and an 8192-token output limit per request.

The summarizer has no access to acquired code, search ledgers, or validation/test performance. Existing family skills are reused rather than regenerated for each program-acquisition run. Formatting failures retain the last usable fused document; absent trajectories or a failure to produce a usable skill leave the base route. Unlike programs, skills undergo no accuracy or gain acceptance test.

A training-side route file records the family, selected program, skill path, and fallback mode. Under \texttt{fallback}, a deployment-eligible program has priority; its out-of-scope inputs use the unmodified executor. A family with no eligible program uses its text skill when available, otherwise the base agent. The driver reads the frozen route. \texttt{always} additionally applies skills outside program scope; \texttt{off} disables skills for core ablations. Transfer with a Flash-Lite-derived skill remains a transfer of that source's experience even when the deployment agent is Gemini-3.1-Pro; independently acquired Gemini-3.1-Pro artifacts are a separate configuration.

\subsection{Policy Gain Decomposition}
\label{app:policy_gain}

For an evaluation set of $N$ examples, let $b_i$ and $h_i$ denote the measured single-run accuracies of the base and skill-guided agents, respectively. Let $c_i$ denote program correctness, $s_i$ indicate direct program adoption, and $t_i=\ind[r_{f(i)}=\mathrm{skill}]$ indicate the text-skill route. Since $s_i t_i=0$, the policy score is $p_i=s_i c_i+t_i h_i+(1-s_i-t_i)b_i$. Its mean gain over the base agent is
\begin{equation}
 \Delta=\underbrace{\frac{1}{N}\sum_i s_i(c_i-b_i)}_{\text{program contribution}}
 +\underbrace{\frac{1}{N}\sum_i t_i(h_i-b_i)}_{\text{skill-fallback contribution}}.
 \label{eq:gain}
\end{equation}
The program term separates into correction mass $N^{-1}\sum_i s_i c_i(1-b_i)$ minus harmful takeover mass $N^{-1}\sum_i s_i(1-c_i)b_i$. This decomposition tracks what direct execution repairs and what it overrides incorrectly. The second term measures text-skill contribution on a disjoint set of inputs.

\section{Sampling Analysis}
\label{app:diagnosis}

\input{sections/sampling_analysis}

\section{Evaluation Accounting and Controls}
\label{app:protocol}

\subsection{Dataset Partitions}
\label{app:development}

Table~\ref{tab:data} lists the training and evaluation partitions. Program writing, scope fitting, and deployment checks use training data only.

\begin{table}[ht]
\caption{Training and evaluation partitions.}
\label{tab:data}
\begin{center}
\begin{tabular}{@{}lrrr@{}}
\toprule
Dataset & Families & Train & Test\\
\midrule
TIR-Bench & 12 & 369 & 746\\
MIRA & 20 & 183 & 363\\
GameQA & 17 & 259 & 514\\
PuzzleVQA & 20 & 663 & 1337\\
\bottomrule
\end{tabular}
\end{center}
\end{table}

\subsection{Model Roles and External Baseline Adaptations}
\label{app:baselines}

Gemini-3.1-Pro authors the external artifacts, which remain fixed for deployment with Gemini-3.1-Pro and Gemini-3.1-Flash-Lite. XSkill, SkillOpt, and Trace2Skill have tool access; GEPA and TextGrad use tool-free inference. Direct answering and the tool agent provide their respective controls.

We extend Trace2Skill with a visual component for image-based tasks. Trace2Skill, SkillOpt, XSkill, and \method{} use the same tool suite.
Appendix~\ref{app:main_accounting} gives aggregation and evaluation details.

\subsection{Frozen Transfer versus Deployment Calibration}
\label{app:transfer_protocol}

Gemini-3.1-Pro and Gemini-3.1-Flash-Lite retain their own training-calibrated routes. DeepSeek and GPT reuse the frozen FL route without target-specific training checks. Programs, scopes, adapters, and stored readings are shared; the target agent executes text skills and the base-agent branch. Additional Pro-policy evaluations with Flash-Lite and DeepSeek appear in Appendix~\ref{app:main_accounting}.

Replacing the deployment agent with executor $e$ changes the agent branches of Equation~\ref{eq:policy}. Equation~\ref{eq:gain} uses that executor's $b_i^{(e)}$ and $h_i^{(e)}$, while the programs, scopes, and internal dependencies remain fixed. With identical stored readings and routes, the direct takeover set and program answers are shared by construction. Their utility remains relative to the target baseline.

A deployment-calibration variant retains $P_f$, $g_f$, and skills but recomputes the family route using target-executor training statistics. The training-side check requires at least five paired in-scope training readings and a positive one-sided 95\% bootstrap lower bound for mean program-minus-executor correctness. A rejected program uses a Pro-trajectory text skill when available. This variant consumes target-executor training experience and changes the route.

\input{sections/main_accounting}

\input{sections/tir_results}
\input{sections/mira_results}

\input{sections/family_results}

\subsection{Illustrative Program Executions}
\label{app:cases}
\begin{figure}[!t]
\centering
\includegraphics[width=\linewidth]{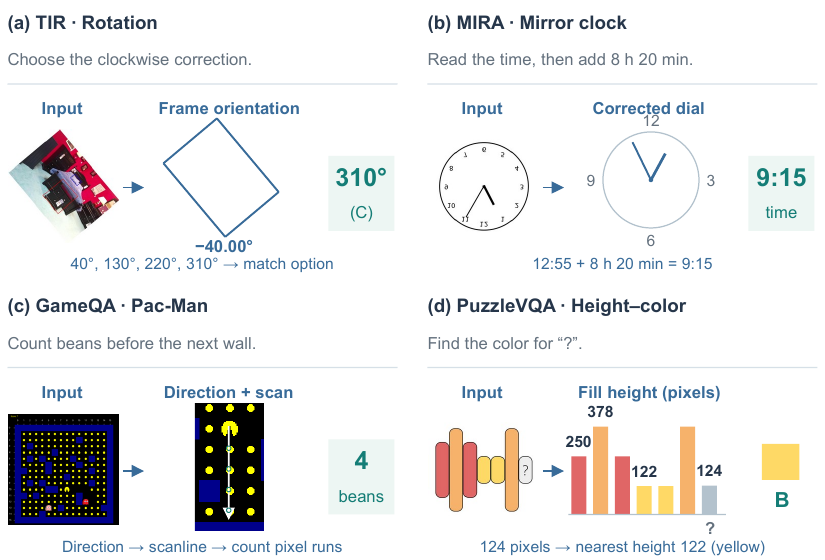}
\caption{\textbf{Additional executable procedures.} (a) Minimum-area frame orientation yields four angle candidates; the answer adapter matches the provided options to select C ($310^\circ$ clockwise). (b) Measured hand directions, mirror/dial correction, and time arithmetic yield 9:15; the central dial depicts corrected angles schematically. (c) Centroid offset determines Pac-Man's direction; the enlarged scanline counts four beans before a wall. (d) Component fill heights map the query bar to yellow (B). All four examples use programs without model calls.}
\label{fig:cases_appendix}
\end{figure}

Figure~\ref{fig:cases} in the main text shows one selected successful execution per dataset. Figure~\ref{fig:cases_appendix} adds one complementary operation per dataset: TIR rotation, MIRA mirror-clock reading, GameQA Pac-Man scanning, and PuzzleVQA height--color lookup. All four additional programs make no model calls.

\subsubsection{From Skill Guidance to Executed Decisions}
\label{app:case_mechanisms}

\paragraph{Completing the suggested computation.}
On the illustrated jigsaw instance, XSkill computes masked boundary MSE and prints the three best right and bottom neighbors for each piece. It then specifies complete layouts directly in plotting calls, without a layout optimizer. Trace2Skill first encounters a boundary-array shape mismatch, resizes the pieces, and prints 30 candidate edge pairs. Its remaining calls likewise plot manually specified layouts. SkillOpt's six calls extract and plot pieces but never compute the pairwise affinities or assembly search described in its skill. The three final answers match their last proposed layouts and are incorrect. These traces locate failures before a correct complete reconstruction has been obtained. In the maze case, Trace2Skill and SkillOpt use no tools and report six turns along a route that crosses an obstacle. XSkill instead extracts the grid, searches direction states, and verifies a path with 17 direction segments, yielding the correct 16 turns.

\paragraph{Selecting a measurable relation.}
SkillOpt's mirror-pattern skill correctly invokes chirality, but its tool-free response selects A from an incorrect reading of the petal sequence. XSkill and Trace2Skill compute SIFT correspondences and affine transformations, identifying C through a negative determinant. The acquired program uses reflected-minus-original ORB match counts of $-26$, $-34$, $+20$, and $-24$ for A--D, respectively, and returns C. On the area--number case, Trace2Skill retains both quantitative size comparison and opposite-number arithmetic as possible strategies, but executes no measurement and extrapolates an opposite-pair sum of eight to the queried pair, incorrectly answering two. XSkill and SkillOpt measure circle radii and correctly answer A. Our program fixes the area relation: the queried component has 16,521 foreground pixels, versus 16,467 for the component labeled six. One model call transcribes characters, and numerical matching returns six (A). These contrasts distinguish knowing a relevant principle from selecting and implementing the right test on an input.

\paragraph{Connection to the deployment comparisons.}
On jigsaw, Gemini-3.1-Pro family accuracy is 2.5\% for XSkill, 7.5\% for Trace2Skill, 10.0\% for SkillOpt, and 95.0\% for \method{} (Table~\ref{tab:family_tirbench_pro}). Holding the acquired program fixed, textual description, callable code, and direct execution yield 33.8\%, 72.5\%, and 95.0\%, respectively. On mirror patterns, the corresponding scores are 80\%, 90\%, and 100\%; reading injection yields 85\% on the same 20 questions. These controls complement the selected traces by varying how the same acquired experience is deployed. For the area--number family, XSkill, SkillOpt, and TextGrad also attain 100\% with Gemini-3.1-Pro.

\subsubsection{Training Feedback and Retained Implementations}
\label{app:jigsaw_acquisition}

The solver in Figure~\ref{fig:cases}(a) is the frozen sixth-round candidate from the acquisition run used by the main experiments. Table~\ref{tab:jigsaw_revisions} traces selected candidates on the same 40 training puzzles. The initial program already used boundary continuity, excluded overlaid number regions from edge comparisons, and assembled pieces with spatial constraints. Failure feedback led the second candidate to try Prim-style assembly from every starting piece, recovering outputs where the first search became stuck. The sixth candidate generated assemblies from several color and gradient cost variants and fixed-seed perturbations, then ranked unique complete grids by the root-mean-square of their boundary costs. This ranks the available assemblies under a shared criterion and increases the penalty for a large discontinuity. It is a reusable heuristic for this task family.

\begin{table}[htbp]
\centering
\caption{\textbf{Selected candidates in the deployed jigsaw solver's acquisition history.} Correct counts use all 40 training puzzles before scope filtering; missing outputs count as incorrect. These are candidates within one historical acquisition run, separate from the controlled acquisition ablations.}
\label{tab:jigsaw_revisions}
\begin{tabular}{@{}cp{0.69\linewidth}r@{}}
\toprule
Round & Revision & Correct / 40\\
\midrule
1 & Boundary matching with constrained Kruskal/Prim-style assembly. & 35\\
2 & Run Prim-style assembly from every starting piece and compare complete grids. & 37\\
4 & Jointly trim two boundary pixels and replace RMSE costs with MSE; rejected after regression. & 18\\
6 & Combine multiple cost variants and fixed-seed perturbations; rank complete grids by RMS boundary cost. & 39\\
\bottomrule
\end{tabular}
\end{table}

The first candidate produced 35 valid answers, all correct; the second produced 40, of which 37 were correct. The fourth candidate's joint preprocessing and scoring change reduced correctness to 18/40 and was rejected. The sixth retained all 35 initially correct answers and solved four additional puzzles. Later candidates did not improve the selection score, so the sixth was frozen. The resulting program uses no model calls and answers 76/80 evaluation puzzles correctly. This trajectory illustrates refinement of an executable procedure through cross-instance feedback, including detection of regressions.

%% file: sections/sampling_analysis.tex
\subsection{Family-Level Performance and Observed Sampling Recovery}
\label{app:sampling}

\paragraph{Separate training and evaluation attempts.}
The program writer uses $K_{\mathrm{tr}}$ training rollouts only through their outcome statistics, together with labeled instances and candidate execution feedback. This TIR analysis uses $K_{\mathrm{te}}=4$ stored Pro base-agent attempts on each of 746 evaluation questions, sampled at temperature 0.6. Policy estimates reuse frozen program readings and per-task skill/base means; skill repetitions are four except word search (one). Neither evaluation labels nor these evaluation attempts enter program writing, scope fitting, route selection, or skill construction.

Let $u_{ij}\in\{0,1\}$ indicate correctness of evaluation attempt $j$ on task $i$, and define $C_i=\sum_{j=1}^{K_{\mathrm{te}}}u_{ij}$ and $O_i=\ind[C_i>0]$. On the same $N$ tasks,
\begin{equation}
 \widehat{\mathrm{pass@1}}_0=\frac{1}{N}\sum_i\frac{C_i}{K_{\mathrm{te}}},\qquad
 \widehat{\mathrm{pass@}K_{\mathrm{te}}}_0=\frac{1}{N}\sum_i O_i.
 \label{eq:sampling_reference}
\end{equation}
The second statistic uses oracle correctness to identify whether any of the four observed answers succeeds. Every question has four recorded base attempts; the family comparisons and strata below use this complete pool.

\paragraph{Comparison on shared task identifiers.}
The base attempts and full VisualCoding policy are paired on the same 746 questions, including failed acquisition, rejected inputs, and text/base fallback. The frozen programs retain their acquisition references (Appendix~\ref{app:tir_results}). Agent answers use the Pro judge, as documented in Appendix~\ref{app:scoring}.

\paragraph{Three observed strata.}
Partition tasks into $\mathcal S_0=\{i:C_i=0\}$, $\mathcal S_{\mathrm{part}}=\{i:0<C_i<K_{\mathrm{te}}\}$, and $\mathcal S_{\mathrm{all}}=\{i:C_i=K_{\mathrm{te}}\}$. Per-task single-run accuracy $\hat p_i$ averages the stored policy outcomes without best-output selection. Table~\ref{tab:sampling_strata} reports the task count and task-weighted mean accuracy in each stratum, along with complete-panel accuracy.

\begin{table}[htbp]
\caption{\textbf{Success stratified by four observed Pro attempts on TIR.} Entries are pass@1 (\%) on common baseline-defined strata. CORE disables text skills but keeps the same programs, scopes, and family eligibility; the full policy retains skill fallback. The tool-agent row uses the reference pool that defines the strata.}
\label{tab:sampling_strata}
\begin{center}
\begin{tabular}{@{}lcccc@{}}
\toprule
Method & $C_i=0$ & $0<C_i<K_{\mathrm{te}}$ & $C_i=K_{\mathrm{te}}$ & Full panel\\
\midrule
Task count & 194 & 211 & 341 & 746\\
\midrule
Tool agent (reference pool) & 0.00 & 52.25 & 100.00 & 60.49\\
VisualCoding CORE & 41.75 & 62.80 & 100.00 & 74.33\\
VisualCoding full & 50.90 & 65.76 & 97.21 & 76.27\\
\bottomrule
\end{tabular}
\end{center}
\end{table}

\begin{figure}[!t]
\centering
\includegraphics[width=\linewidth]{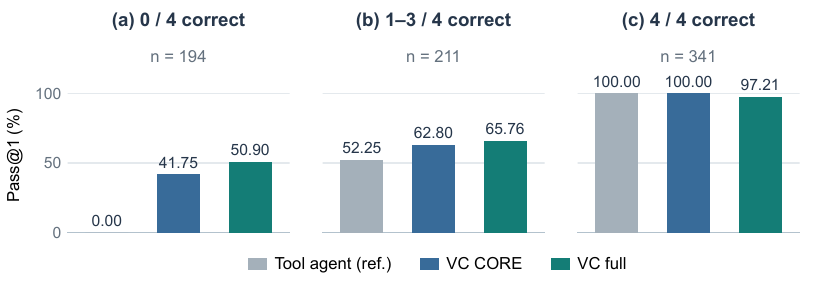}
\caption{\textbf{Single-run accuracy across observed sampling strata.} TIR-Bench tasks are grouped by the number of correct answers among four stored Pro tool-agent attempts. Bars reproduce Table~\ref{tab:sampling_strata}; CORE retains programs and scopes with text fallback disabled. Grouping fixes the reference rates at 0\% and 100\% in the extreme strata.}
\label{fig:sampling_analysis}
\end{figure}

Figure~\ref{fig:sampling_analysis} visualizes these three policies under the common baseline-defined strata.

\paragraph{TIR recovery and retained success.}
Among $\mathcal S_0$'s 194 tasks, 88 receive program takeovers and 81 are correct: 69 jigsaw, four maze, seven OCR, and one rotation question. These are successful program answers on questions for which all four recorded base-agent attempts fail. Of the remaining tasks in this stratum, 105 receive skills, contributing 17.75 summed per-task success estimates, and one receives base fallback. The fractional skill total sums estimated probabilities rather than counting distinct solved tasks. On the 211 intermittently solved tasks, CORE attains 62.80\% single-run accuracy, compared with 52.25\% for the base agent. In $\mathcal S_{\mathrm{all}}$, CORE preserves all observed successes; the full policy attains 97.21\%, with its 2.79-point conditional loss arising on skill routes.

\paragraph{Family-level comparison.}
Table~\ref{tab:tir_sampling_families} reports all twelve families. Full-policy estimated pass@1 exceeds the base agent's observed pass@4 in four: jigsaw (95.00\% versus 11.25\%), maze (100.00\% versus 95.00\%), OCR (87.50\% versus 72.50\%), and rotation game (100.00\% versus 98.00\%). Together with the per-question recovery counts, these comparisons show successful program answers beyond those observed in the four recorded baseline attempts.

\begin{table}[htbp]
\caption{\textbf{Family-level Pro sampling reference and full-policy accuracy on TIR.} Percentages; all families are reported. OCR uses stored model-dependent program readings. Other rows can include program, skill, or base routes; these are complete-family scores, not in-scope precision.}
\label{tab:tir_sampling_families}
\begin{center}
\begin{tabular}{@{}lrrrr@{}}
\toprule
Family & $N$ & Base @1 & Base @4 & Full @1\\
\midrule
Color & 67 & 71.64 & 85.07 & 72.39\\
Contrast & 34 & 63.24 & 79.41 & 63.97\\
Instrument & 54 & 22.69 & 37.04 & 25.00\\
Jigsaw & 80 & 4.38 & 11.25 & 95.00\\
Math & 80 & 74.69 & 83.75 & 75.62\\
Maze & 80 & 88.12 & 95.00 & 100.00\\
OCR & 40 & 55.62 & 72.50 & 87.50\\
RefCOCO & 80 & 44.06 & 70.00 & 58.12\\
Rotation game & 50 & 89.50 & 98.00 & 100.00\\
Symbolic & 34 & 78.68 & 94.12 & 88.24\\
Visual search & 80 & 66.88 & 81.25 & 67.81\\
Word search & 67 & 79.48 & 97.01 & 79.10\\
\bottomrule
\end{tabular}
\end{center}
\end{table}

\paragraph{Interpretation of the sampling reference.}
The four recorded attempts provide a finite sampling reference rather than a capability upper bound. Additional attempts may move questions out of $\mathcal S_0$.

%% file: sections/main_accounting.tex
\subsection{Four-Benchmark Results: Aggregation and Evaluation}
\label{app:main_accounting}

\paragraph{Aggregation.}
The four-benchmark suite comprises 69 task families, with 1,474 training questions and 2,960 test questions. Dataset accuracy is $\sum_f n_f a_f/\sum_f n_f$, where $n_f$ is the number of test questions and $a_f$ is the accuracy in family $f$. Avg. is the unweighted mean of the four dataset accuracies. Tables report accuracy to one decimal percentage point.

\paragraph{Model configuration and transfer.}
Pro-learned external artifacts remain fixed when deployed with Flash-Lite. Our Pro and Flash-Lite deployments use their respective training-calibrated family routes. Pro uses low thinking; Flash-Lite and DeepSeek disable thinking. DeepSeek and GPT reuse the frozen FL source route across all four datasets, without target-model training rollouts or deployment checks. Each target baseline and skill response is sampled once.

\paragraph{Answer scoring.}
\label{app:scoring}
For agent answers, the model judge compares the prediction with a supplied reference answer. In the main comparisons, Gemini 3.1 Pro experiments use Gemini 3.1 Pro as the judge; Flash-Lite, DeepSeek, and GPT experiments use Gemini 3.1 Flash-Lite. Because grading is a reference-based answer comparison, we use the lighter Flash-Lite judge for these subsequent evaluations.

\paragraph{Reported estimates and sampling.}
Main comparisons and ablations report point estimates. Acquisition ablations evaluate re-acquired family programs under a shared deployment protocol; deployment ablations use frozen artifacts with one recorded agent response per intervention task. The main TIR comparison and ablations share the fixed first rollout for base-agent and text-skill branches, giving a Pro tool-agent reference of 62.87\%. 

\paragraph{Results and interpretation.}
VisualCoding's four-dataset averages are 76.11\% (Pro) and 68.57\% (Flash-Lite), gains of 9.33 and 25.14 points over their own tool controls. Among the evaluated external methods, XSkill gives the highest averages, 71.67\% and 53.78\%; individual datasets can favor another external method. SkillOpt's four-dataset averages are 70.43\% (Pro) and 53.33\% (Flash-Lite).

\paragraph{Source-policy sensitivity.}
Here, a deployment policy specifies whether each family uses its retained program or a fallback skill. Table~\ref{tab:fl_policy_sensitivity} compares Flash-Lite under the frozen Pro policy, its own calibrated policy, and the Pro policy with the additional executor-specific deployment check removed. Programs, scopes, and retained skills remain fixed. The checked Pro and FL policies enable programs in 27 and 40 of the 69 families, respectively. For the 13 families that switch from programs to text skills under the Pro policy, we collect 649 Flash-Lite skill-guided responses; unchanged outputs and frozen program readings are reused. 

The frozen Pro policy improves Flash-Lite on all four benchmarks, raising equal-dataset mean accuracy from 43.42\% to 61.92\%. Removing the executor-specific check raises this mean to 65.15\%, while Flash-Lite-specific calibration reaches 68.57\%. These results suggest that a check calibrated against Pro can exclude programs useful to a weaker executor. The no-check condition retains source eligibility and program scope, so its gain concerns family-level program selection.

\begin{table}[t]
\caption{\textbf{Flash-Lite sensitivity to the family deployment policy.} Accuracy (\%); every column uses Flash-Lite as the executor. Pro and FL denote the executor used for deployment calibration; no check removes the additional executor-specific check from the Pro configuration. The final column reports Pro-policy minus FL-policy accuracy in percentage points, with paired task-bootstrap 95\% intervals from 4,000 resamples for each dataset and the pooled sample. Avg. equally weights datasets; Pooled weights all 2,960 questions.}
\label{tab:fl_policy_sensitivity}
\centering\small\setlength{\tabcolsep}{4pt}
\begin{tabular}{@{}lrrrrr@{}}
\toprule
Dataset ($n$) & \shortstack[r]{Tool\\agent} & \shortstack[r]{Pro\\policy} & \shortstack[r]{Pro\\no check} & \shortstack[r]{FL\\policy} & \shortstack[r]{Pro $-$ FL\\{[95\% CI]}}\\
\midrule
TIR-Bench (746) & 39.0 & 61.5 & 65.0 & 68.5 & $-7.0\;[-9.4, -4.7]$\\
MIRA (363) & 20.9 & 36.4 & 38.3 & 41.6 & $-5.2\;[-8.0, -3.0]$\\
GameQA (514) & 30.4 & 52.7 & 58.6 & 65.4 & $-12.6\;[-16.0, -9.5]$\\
PuzzleVQA (1,337) & 83.4 & 97.1 & 98.7 & 98.8 & $-1.7\;[-2.5, -0.9]$\\
\midrule
Avg. (datasets) & 43.4 & 61.9 & 65.1 & 68.6 & $-6.6$\\
Pooled (2,960) & 55.3 & 73.0 & 75.8 & 78.3 & $-5.4\;[-6.3, -4.5]$\\
\bottomrule
\end{tabular}
\end{table}

DeepSeek's mean is 69.76\% with the Pro policy and 71.66\% with the FL policy, versus 60.74\% for its tool agent. Source-policy effects vary: the Pro policy is better on TIR (76.54\% versus 75.60\%), whereas the FL policy is better on GameQA (68.87\% versus 63.23\%), MIRA (44.35\% versus 41.87\%), and PuzzleVQA (97.83\% versus 97.38\%).

%% file: sections/tir_results.tex
\subsection{TIR-Bench Ablations}
\label{app:tir_results}

\paragraph{Panel and acquisition.}
TIR-Bench contains 12 families, 746 test questions, and 369 training questions. The paired test baseline uses the same fixed first Pro rollout per question as Table~\ref{tab:main}; agent answers use the Pro judge (Appendix~\ref{app:scoring}). The acquisition-feedback comparison evaluates Full execution feedback (FULLC), Independent searc (IID), SCALAR and no scope on jigsaw, RefCOCO, word search, symbolic, and maze. Within each family, these arms share initial facts, training references, candidate budgets, and selection rules. FULLC uses earlier candidates and their execution feedback; IID proposes candidates independently while retaining the same execution and selection process. Pro training references use four rollouts per question except for jigsaw, which uses one. Jigsaw has 18 candidate slots; the other families have 36.

For both benchmarks, complete-panel evaluation combines the re-acquired family programs with the main experiment's fixed single-rollout outputs on unaffected families. Base-agent fallback uses the fixed first rollout on each question; unchanged skill branches use their fixed first skill-guided rollout. On re-acquired families, eligibility requires source acceptance and at least 50\% training coverage; text skills and the additional executor-specific check are disabled, and failed acquisitions revert to the base agent. SCALAR retains the best code and aggregate execution outcomes but omits per-example feedback and error-selected images. Its complete-panel accuracy is 69.7\% on TIR. Appendix~\ref{app:family_ablations} reports the complete-family acquisition and deployment results.

\paragraph{Shared representation and continuation protocols.}
The representation and answer-continuation comparisons use the frozen main-experiment programs and their original acquisition references; the shared protocols below apply to both benchmarks. REP-TEXT supplies a procedure translated by Pro from the frozen program, checked for omissions and interface consistency on five training questions per family before freezing. These procedures are distinct from the system's trajectory-derived fallback skills. REP-CODE supplies the complete source and a callable implementation that the agent may inspect, call, modify, or ignore. Both arms retain coding access and share the primitive library, task interface, answer adapter, original-image access, and a budget of 30 turns and 65,536 tokens per task.

DEC-DIRECT returns the answer from the frozen program reading. DEC-INJECT supplies Pro with the question, image, computed value, permitted diagnostics, applicability status, and candidate answer for continuation without tools or additional perception calls. Readings, scopes, and prompts remain fixed. Each agent intervention has one rollout per question and produces the final answer.

TIR's representation intervention replaces 228 deployed takeovers: jigsaw 80, maze 80, rotation 50, and symbolic 18. OCR's 40 takeovers remain direct because its model interface is unavailable in the agent kernel. The answer-continuation comparison includes these OCR takeovers, affecting 268 questions. All other system outputs stay fixed. The text-skill fallback control appears in Table~\ref{tab:deployment_controls}.

%% file: sections/mira_results.tex
\subsection{MIRA Ablations}
\label{app:mira_results}

\paragraph{Acquisition protocol.}
MIRA contains 20 families and 363 test questions. The acquisition comparison selects the five pure-pixel program families with the lowest Pro training pass@1 using training-side information: billiards, puzzle, mirror pattern, trailer cubes count, and mirror clock. Acquisition uses 36 candidate slots and 7--10 training examples per family. The writer and four-rollout training reference are Pro; test fallback uses one stored Pro rollout. Other acquisition settings follow Appendix~\ref{app:tir_results}. SCALAR's complete-panel accuracy is 46.5\%. Deployment interventions use frozen programs acquired with Flash-Lite training references, separately from the Pro-referenced acquisitions in the feedback comparison.

\paragraph{Deployment interventions.}
The representation and answer-continuation comparisons follow the shared protocol and use the Pro judge. They replace 71 deployed takeovers: 14 billiards, 20 mirror clock, 20 mirror pattern, and 17 puzzle. All other outputs remain fixed, including the original direct outputs of model-read families. In the CODE condition, the agent calls the supplied mirror-clock program on 10\% of questions.

Table~\ref{tab:deployment_controls} isolates text-skill fallback on both benchmarks. The no-text-fallback condition replaces the text-skill branch with the base agent while keeping program routes and their out-of-scope fallback unchanged. Both conditions use the same fixed first base-agent and skill-guided rollouts as the main comparison. Text-skill fallback adds 0.3 percentage points on TIR and 0.8 on MIRA under this shared evaluation protocol.

\begin{table}[htbp]
\caption{\textbf{Text-skill fallback control.} Complete-panel accuracy (\%) with Pro on the same TIR-Bench and MIRA panels as Table~\ref{tab:ablations}.}
\label{tab:deployment_controls}
\centering
\begin{tabular}{@{}lcc@{}}
\toprule
Variant & TIR-Bench & MIRA\\
\midrule
Full method & 76.3 & 47.1\\
No text fallback & 76.0 & 46.3\\
\bottomrule
\end{tabular}
\end{table}

%% file: sections/family_results.tex
\section{Complete Family-Level Results}

\label{app:family_results}

All 69 families in the four-benchmark main suite are included below, in the same benchmark order as Tables~\ref{tab:main} and~\ref{tab:transfer}. Entries are complete-family accuracy (\%), including program rejection and fallback; $n$ counts test questions. Dataset totals are task-weighted, not averages of family percentages. Unrounded family scores reproduce every corresponding main-table dataset entry. Family identifiers omit the dataset prefixes \texttt{mr\_}, \texttt{gq\_}, and \texttt{pz\_}; underscores are displayed as spaces.

\subsection{Main Comparison and Cross-Executor Reuse}

Dir. denotes tool-free direct answering; Tool the tool agent; XS, SO, T2S, GEPA, and TG denote XSkill, SkillOpt, Trace2Skill, GEPA, and TextGrad; VC denotes \method{}. GEPA and TG are tool-free adaptations. Pro and Flash-Lite retain their respective training-calibrated routes. DeepSeek and GPT reuse the FL route and stored program readings.

Bold/underline mark best/second-best distinct displayed accuracies within each executor (ties included). Small green/red numbers are gains/losses in percentage points: GEPA/TextGrad versus Direct; XSkill/Trace2Skill/SkillOpt/VC versus Tool agent. In these dense family tables the change is placed below the accuracy; shaded cells identify VC. Rankings are within each family and executor, with dataset totals ranked separately.

\begingroup
\small\setlength{\tabcolsep}{3pt}
\begin{longtable}{@{}lrrrrrrrrr@{}}
\caption{\textbf{TIR-Bench: Gemini 3.1 Pro.} Complete-family accuracy (\%).}\label{tab:family_tirbench_pro}\\
\toprule
Family & $n$ & Dir. & GEPA & TG & Tool & XS & T2S & SO & \cellcolor{methodshade}VC\\\midrule
\endfirsthead
\multicolumn{10}{l}{\small\tablename~\thetable{} (continued)}\\
\toprule
Family & $n$ & Dir. & GEPA & TG & Tool & XS & T2S & SO & \cellcolor{methodshade}VC\\\midrule
\endhead
\midrule\multicolumn{10}{r}{Continued on next page}\\
\endfoot
\bottomrule\endlastfoot
color & 67 & 50.7 & \shortstack[r]{59.7\\{\scriptsize \textcolor{gaincolor}{+9.0}}} & \shortstack[r]{52.2\\{\scriptsize \textcolor{gaincolor}{+1.5}}} & \underline{74.6} & \shortstack[r]{\underline{74.6}\\{\scriptsize \textcolor{neutralcolor}{0.0}}} & \shortstack[r]{\textbf{77.6}\\{\scriptsize \textcolor{gaincolor}{+3.0}}} & \shortstack[r]{73.1\\{\scriptsize \textcolor{losscolor}{-1.5}}} & \cellcolor{methodshade}\shortstack[r]{70.1\\{\scriptsize \textcolor{losscolor}{-4.5}}}\\
contrast & 34 & 64.7 & \shortstack[r]{47.1\\{\scriptsize \textcolor{losscolor}{-17.6}}} & \shortstack[r]{58.8\\{\scriptsize \textcolor{losscolor}{-5.9}}} & 64.7 & \shortstack[r]{\textbf{73.5}\\{\scriptsize \textcolor{gaincolor}{+8.8}}} & \shortstack[r]{64.7\\{\scriptsize \textcolor{neutralcolor}{0.0}}} & \shortstack[r]{\underline{67.7}\\{\scriptsize \textcolor{gaincolor}{+2.9}}} & \cellcolor{methodshade}\shortstack[r]{52.9\\{\scriptsize \textcolor{losscolor}{-11.8}}}\\
instrument & 54 & 24.1 & \shortstack[r]{\underline{25.9}\\{\scriptsize \textcolor{gaincolor}{+1.9}}} & \shortstack[r]{24.1\\{\scriptsize \textcolor{neutralcolor}{0.0}}} & 24.1 & \shortstack[r]{\underline{25.9}\\{\scriptsize \textcolor{gaincolor}{+1.9}}} & \shortstack[r]{24.1\\{\scriptsize \textcolor{neutralcolor}{0.0}}} & \shortstack[r]{\textbf{31.5}\\{\scriptsize \textcolor{gaincolor}{+7.4}}} & \cellcolor{methodshade}\shortstack[r]{\underline{25.9}\\{\scriptsize \textcolor{gaincolor}{+1.9}}}\\
jigsaw & 80 & 1.2 & \shortstack[r]{3.8\\{\scriptsize \textcolor{gaincolor}{+2.5}}} & \shortstack[r]{5.0\\{\scriptsize \textcolor{gaincolor}{+3.8}}} & 5.0 & \shortstack[r]{2.5\\{\scriptsize \textcolor{losscolor}{-2.5}}} & \shortstack[r]{7.5\\{\scriptsize \textcolor{gaincolor}{+2.5}}} & \shortstack[r]{\underline{10.0}\\{\scriptsize \textcolor{gaincolor}{+5.0}}} & \cellcolor{methodshade}\shortstack[r]{\textbf{95.0}\\{\scriptsize \textcolor{gaincolor}{+90.0}}}\\
\end{longtable}
\endgroup

\begingroup
\small\setlength{\tabcolsep}{3pt}
\begin{longtable}{@{}lrrrrrrrrr@{}}
\caption{\textbf{TIR-Bench: Gemini 3.1 Flash-Lite.} Complete-family accuracy (\%).}\label{tab:family_tirbench_flashlite}\\
\toprule
Family & $n$ & Dir. & GEPA & TG & Tool & XS & T2S & SO & \cellcolor{methodshade}VC\\\midrule
\endfirsthead
\multicolumn{10}{l}{\small\tablename~\thetable{} (continued)}\\
\toprule
Family & $n$ & Dir. & GEPA & TG & Tool & XS & T2S & SO & \cellcolor{methodshade}VC\\\midrule
\endhead
\midrule\multicolumn{10}{r}{Continued on next page}\\
\endfoot
\bottomrule\endlastfoot
color & 67 & 41.8 & \shortstack[r]{46.3\\{\scriptsize \textcolor{gaincolor}{+4.5}}} & \shortstack[r]{52.2\\{\scriptsize \textcolor{gaincolor}{+10.4}}} & 61.2 & \shortstack[r]{64.2\\{\scriptsize \textcolor{gaincolor}{+3.0}}} & \shortstack[r]{\underline{68.7}\\{\scriptsize \textcolor{gaincolor}{+7.5}}} & \shortstack[r]{\textbf{73.1}\\{\scriptsize \textcolor{gaincolor}{+11.9}}} & \cellcolor{methodshade}\shortstack[r]{56.7\\{\scriptsize \textcolor{losscolor}{-4.5}}}\\
contrast & 34 & 41.2 & \shortstack[r]{44.1\\{\scriptsize \textcolor{gaincolor}{+2.9}}} & \shortstack[r]{47.1\\{\scriptsize \textcolor{gaincolor}{+5.9}}} & 55.9 & \shortstack[r]{\underline{61.8}\\{\scriptsize \textcolor{gaincolor}{+5.9}}} & \shortstack[r]{\textbf{64.7}\\{\scriptsize \textcolor{gaincolor}{+8.8}}} & \shortstack[r]{58.8\\{\scriptsize \textcolor{gaincolor}{+2.9}}} & \cellcolor{methodshade}\shortstack[r]{55.9\\{\scriptsize \textcolor{neutralcolor}{0.0}}}\\
instrument & 54 & 14.8 & \shortstack[r]{20.4\\{\scriptsize \textcolor{gaincolor}{+5.6}}} & \shortstack[r]{14.8\\{\scriptsize \textcolor{neutralcolor}{0.0}}} & 13.0 & \shortstack[r]{\underline{24.1}\\{\scriptsize \textcolor{gaincolor}{+11.1}}} & \shortstack[r]{\underline{24.1}\\{\scriptsize \textcolor{gaincolor}{+11.1}}} & \shortstack[r]{\textbf{27.8}\\{\scriptsize \textcolor{gaincolor}{+14.8}}} & \cellcolor{methodshade}\shortstack[r]{16.7\\{\scriptsize \textcolor{gaincolor}{+3.7}}}\\
jigsaw & 80 & \underline{0.0} & \shortstack[r]{\underline{0.0}\\{\scriptsize \textcolor{neutralcolor}{0.0}}} & \shortstack[r]{\underline{0.0}\\{\scriptsize \textcolor{neutralcolor}{0.0}}} & \underline{0.0} & \shortstack[r]{\underline{0.0}\\{\scriptsize \textcolor{neutralcolor}{0.0}}} & \shortstack[r]{\underline{0.0}\\{\scriptsize \textcolor{neutralcolor}{0.0}}} & \shortstack[r]{\underline{0.0}\\{\scriptsize \textcolor{neutralcolor}{0.0}}} & \cellcolor{methodshade}\shortstack[r]{\textbf{95.0}\\{\scriptsize \textcolor{gaincolor}{+95.0}}}\\
\end{longtable}
\endgroup

\begingroup
\small\setlength{\tabcolsep}{3pt}
\begin{longtable}{@{}lrrrrr@{}}
\caption{\textbf{TIR-Bench: DeepSeek and GPT with frozen FL routes.} Complete-family accuracy (\%).}\label{tab:family_tirbench_transfer}\\
\toprule
Family & $n$ & DS Tool & \cellcolor{methodshade}DS VC & GPT Tool & \cellcolor{methodshade}GPT VC\\\midrule
\endfirsthead
\multicolumn{6}{l}{\small\tablename~\thetable{} (continued)}\\
\toprule
Family & $n$ & DS Tool & \cellcolor{methodshade}DS VC & GPT Tool & \cellcolor{methodshade}GPT VC\\\midrule
\endhead
\midrule\multicolumn{6}{r}{Continued on next page}\\
\endfoot
\bottomrule\endlastfoot
color & 67 & \textbf{74.6} & \cellcolor{methodshade}\shortstack[r]{\underline{64.2}\\{\scriptsize \textcolor{losscolor}{-10.4}}} & \underline{50.7} & \cellcolor{methodshade}\shortstack[r]{\textbf{62.7}\\{\scriptsize \textcolor{gaincolor}{+11.9}}}\\
contrast & 34 & \underline{58.8} & \cellcolor{methodshade}\shortstack[r]{\textbf{61.8}\\{\scriptsize \textcolor{gaincolor}{+2.9}}} & \underline{26.5} & \cellcolor{methodshade}\shortstack[r]{\textbf{32.4}\\{\scriptsize \textcolor{gaincolor}{+5.9}}}\\
instrument & 54 & \underline{14.8} & \cellcolor{methodshade}\shortstack[r]{\textbf{25.9}\\{\scriptsize \textcolor{gaincolor}{+11.1}}} & \underline{16.7} & \cellcolor{methodshade}\shortstack[r]{\textbf{31.5}\\{\scriptsize \textcolor{gaincolor}{+14.8}}}\\
jigsaw & 80 & \underline{5.0} & \cellcolor{methodshade}\shortstack[r]{\textbf{95.0}\\{\scriptsize \textcolor{gaincolor}{+90.0}}} & \underline{1.2} & \cellcolor{methodshade}\shortstack[r]{\textbf{95.0}\\{\scriptsize \textcolor{gaincolor}{+93.8}}}\\
math & 80 & \textbf{81.2} & \cellcolor{methodshade}\shortstack[r]{\underline{77.5}\\{\scriptsize \textcolor{losscolor}{-3.7}}} & \textbf{37.5} & \cellcolor{methodshade}\shortstack[r]{\underline{35.0}\\{\scriptsize \textcolor{losscolor}{-2.5}}}\\
maze & 80 & \underline{68.8} & \cellcolor{methodshade}\shortstack[r]{\textbf{100.0}\\{\scriptsize \textcolor{gaincolor}{+31.2}}} & \underline{57.5} & \cellcolor{methodshade}\shortstack[r]{\textbf{100.0}\\{\scriptsize \textcolor{gaincolor}{+42.5}}}\\
ocr & 40 & \underline{40.0} & \cellcolor{methodshade}\shortstack[r]{\textbf{87.5}\\{\scriptsize \textcolor{gaincolor}{+47.5}}} & \underline{37.5} & \cellcolor{methodshade}\shortstack[r]{\textbf{87.5}\\{\scriptsize \textcolor{gaincolor}{+50.0}}}\\
refcoco & 80 & \underline{53.8} & \cellcolor{methodshade}\shortstack[r]{\textbf{60.0}\\{\scriptsize \textcolor{gaincolor}{+6.2}}} & \underline{38.8} & \cellcolor{methodshade}\shortstack[r]{\textbf{60.0}\\{\scriptsize \textcolor{gaincolor}{+21.2}}}\\
\end{longtable}
\endgroup

\begingroup
\small\setlength{\tabcolsep}{3pt}
\begin{longtable}{@{}lrrrrrrrrr@{}}
\caption{\textbf{MIRA: Gemini 3.1 Pro.} Complete-family accuracy (\%).}\label{tab:family_mira_pro}\\
\toprule
Family & $n$ & Dir. & GEPA & TG & Tool & XS & T2S & SO & \cellcolor{methodshade}VC\\\midrule
\endfirsthead
\multicolumn{10}{l}{\small\tablename~\thetable{} (continued)}\\
\toprule
Family & $n$ & Dir. & GEPA & TG & Tool & XS & T2S & SO & \cellcolor{methodshade}VC\\\midrule
\endhead
\midrule\multicolumn{10}{r}{Continued on next page}\\
\endfoot
\bottomrule\endlastfoot
billiards & 14 & 14.3 & \shortstack[r]{28.6\\{\scriptsize \textcolor{gaincolor}{+14.3}}} & \shortstack[r]{50.0\\{\scriptsize \textcolor{gaincolor}{+35.7}}} & 35.7 & \shortstack[r]{\underline{64.3}\\{\scriptsize \textcolor{gaincolor}{+28.6}}} & \shortstack[r]{50.0\\{\scriptsize \textcolor{gaincolor}{+14.3}}} & \shortstack[r]{50.0\\{\scriptsize \textcolor{gaincolor}{+14.3}}} & \cellcolor{methodshade}\shortstack[r]{\textbf{71.4}\\{\scriptsize \textcolor{gaincolor}{+35.7}}}\\
convex hull & 20 & 55.0 & \shortstack[r]{55.0\\{\scriptsize \textcolor{neutralcolor}{0.0}}} & \shortstack[r]{25.0\\{\scriptsize \textcolor{losscolor}{-30.0}}} & \underline{90.0} & \shortstack[r]{\underline{90.0}\\{\scriptsize \textcolor{neutralcolor}{0.0}}} & \shortstack[r]{\textbf{95.0}\\{\scriptsize \textcolor{gaincolor}{+5.0}}} & \shortstack[r]{\textbf{95.0}\\{\scriptsize \textcolor{gaincolor}{+5.0}}} & \cellcolor{methodshade}\shortstack[r]{\underline{90.0}\\{\scriptsize \textcolor{neutralcolor}{0.0}}}\\
cubes count & 20 & \textbf{5.0} & \shortstack[r]{\textbf{5.0}\\{\scriptsize \textcolor{neutralcolor}{0.0}}} & \shortstack[r]{\underline{0.0}\\{\scriptsize \textcolor{losscolor}{-5.0}}} & \textbf{5.0} & \shortstack[r]{\textbf{5.0}\\{\scriptsize \textcolor{neutralcolor}{0.0}}} & \shortstack[r]{\textbf{5.0}\\{\scriptsize \textcolor{neutralcolor}{0.0}}} & \shortstack[r]{\underline{0.0}\\{\scriptsize \textcolor{losscolor}{-5.0}}} & \cellcolor{methodshade}\shortstack[r]{\textbf{5.0}\\{\scriptsize \textcolor{neutralcolor}{0.0}}}\\
\end{longtable}
\endgroup

\begingroup
\small\setlength{\tabcolsep}{3pt}
\begin{longtable}{@{}lrrrrrrrrr@{}}
\caption{\textbf{MIRA: Gemini 3.1 Flash-Lite.} Complete-family accuracy (\%).}\label{tab:family_mira_flashlite}\\
\toprule
Family & $n$ & Dir. & GEPA & TG & Tool & XS & T2S & SO & \cellcolor{methodshade}VC\\\midrule
\endfirsthead
\multicolumn{10}{l}{\small\tablename~\thetable{} (continued)}\\
\toprule
Family & $n$ & Dir. & GEPA & TG & Tool & XS & T2S & SO & \cellcolor{methodshade}VC\\\midrule
\endhead
\midrule\multicolumn{10}{r}{Continued on next page}\\
\endfoot
\bottomrule\endlastfoot
billiards & 14 & 28.6 & \shortstack[r]{0.0\\{\scriptsize \textcolor{losscolor}{-28.6}}} & \shortstack[r]{28.6\\{\scriptsize \textcolor{neutralcolor}{0.0}}} & 35.7 & \shortstack[r]{\underline{42.9}\\{\scriptsize \textcolor{gaincolor}{+7.1}}} & \shortstack[r]{28.6\\{\scriptsize \textcolor{losscolor}{-7.1}}} & \shortstack[r]{28.6\\{\scriptsize \textcolor{losscolor}{-7.1}}} & \cellcolor{methodshade}\shortstack[r]{\textbf{71.4}\\{\scriptsize \textcolor{gaincolor}{+35.7}}}\\
convex hull & 20 & 15.0 & \shortstack[r]{20.0\\{\scriptsize \textcolor{gaincolor}{+5.0}}} & \shortstack[r]{10.0\\{\scriptsize \textcolor{losscolor}{-5.0}}} & \underline{85.0} & \shortstack[r]{\textbf{90.0}\\{\scriptsize \textcolor{gaincolor}{+5.0}}} & \shortstack[r]{\textbf{90.0}\\{\scriptsize \textcolor{gaincolor}{+5.0}}} & \shortstack[r]{80.0\\{\scriptsize \textcolor{losscolor}{-5.0}}} & \cellcolor{methodshade}\shortstack[r]{\underline{85.0}\\{\scriptsize \textcolor{neutralcolor}{0.0}}}\\
cubes count & 20 & \underline{0.0} & \shortstack[r]{\underline{0.0}\\{\scriptsize \textcolor{neutralcolor}{0.0}}} & \shortstack[r]{\underline{0.0}\\{\scriptsize \textcolor{neutralcolor}{0.0}}} & \underline{0.0} & \shortstack[r]{\textbf{5.0}\\{\scriptsize \textcolor{gaincolor}{+5.0}}} & \shortstack[r]{\underline{0.0}\\{\scriptsize \textcolor{neutralcolor}{0.0}}} & \shortstack[r]{\underline{0.0}\\{\scriptsize \textcolor{neutralcolor}{0.0}}} & \cellcolor{methodshade}\shortstack[r]{\underline{0.0}\\{\scriptsize \textcolor{neutralcolor}{0.0}}}\\
\end{longtable}
\endgroup

\begingroup
\small\setlength{\tabcolsep}{3pt}
\begin{longtable}{@{}lrrrrr@{}}
\caption{\textbf{MIRA: DeepSeek and GPT with frozen FL routes.} Complete-family accuracy (\%).}\label{tab:family_mira_transfer}\\
\toprule
Family & $n$ & DS Tool & \cellcolor{methodshade}DS VC & GPT Tool & \cellcolor{methodshade}GPT VC\\\midrule
\endfirsthead
\multicolumn{6}{l}{\small\tablename~\thetable{} (continued)}\\
\toprule
Family & $n$ & DS Tool & \cellcolor{methodshade}DS VC & GPT Tool & \cellcolor{methodshade}GPT VC\\\midrule
\endhead
\midrule\multicolumn{6}{r}{Continued on next page}\\
\endfoot
\bottomrule\endlastfoot
billiards & 14 & \underline{57.1} & \cellcolor{methodshade}\shortstack[r]{\textbf{71.4}\\{\scriptsize \textcolor{gaincolor}{+14.3}}} & \underline{14.3} & \cellcolor{methodshade}\shortstack[r]{\textbf{71.4}\\{\scriptsize \textcolor{gaincolor}{+57.1}}}\\
convex hull & 20 & \textbf{95.0} & \cellcolor{methodshade}\shortstack[r]{\underline{90.0}\\{\scriptsize \textcolor{losscolor}{-5.0}}} & \textbf{65.0} & \cellcolor{methodshade}\shortstack[r]{\underline{60.0}\\{\scriptsize \textcolor{losscolor}{-5.0}}}\\
cubes count & 20 & \textbf{5.0} & \cellcolor{methodshade}\shortstack[r]{\textbf{5.0}\\{\scriptsize \textcolor{neutralcolor}{0.0}}} & \textbf{0.0} & \cellcolor{methodshade}\shortstack[r]{\textbf{0.0}\\{\scriptsize \textcolor{neutralcolor}{0.0}}}\\
cubes missing & 20 & \textbf{0.0} & \cellcolor{methodshade}\shortstack[r]{\textbf{0.0}\\{\scriptsize \textcolor{neutralcolor}{0.0}}} & \textbf{0.0} & \cellcolor{methodshade}\shortstack[r]{\textbf{0.0}\\{\scriptsize \textcolor{neutralcolor}{0.0}}}\\
defuse a bomb & 17 & \textbf{35.3} & \cellcolor{methodshade}\shortstack[r]{\underline{17.6}\\{\scriptsize \textcolor{losscolor}{-17.6}}} & \underline{23.5} & \cellcolor{methodshade}\shortstack[r]{\textbf{29.4}\\{\scriptsize \textcolor{gaincolor}{+5.9}}}\\
electric charge & 14 & \textbf{71.4} & \cellcolor{methodshade}\shortstack[r]{\textbf{71.4}\\{\scriptsize \textcolor{neutralcolor}{0.0}}} & \underline{50.0} & \cellcolor{methodshade}\shortstack[r]{\textbf{64.3}\\{\scriptsize \textcolor{gaincolor}{+14.3}}}\\
gear rotation & 26 & \textbf{50.0} & \cellcolor{methodshade}\shortstack[r]{\underline{46.2}\\{\scriptsize \textcolor{losscolor}{-3.8}}} & \underline{42.3} & \cellcolor{methodshade}\shortstack[r]{\textbf{50.0}\\{\scriptsize \textcolor{gaincolor}{+7.7}}}\\
\end{longtable}
\endgroup

\begingroup
\small\setlength{\tabcolsep}{3pt}
\begin{longtable}{@{}lrrrrrrrrr@{}}
\caption{\textbf{GameQA-17: Gemini 3.1 Pro.} Complete-family accuracy (\%).}\label{tab:family_gameqa_pro}\\
\toprule
Family & $n$ & Dir. & GEPA & TG & Tool & XS & T2S & SO & \cellcolor{methodshade}VC\\\midrule
\endfirsthead
\multicolumn{10}{l}{\small\tablename~\thetable{} (continued)}\\
\toprule
Family & $n$ & Dir. & GEPA & TG & Tool & XS & T2S & SO & \cellcolor{methodshade}VC\\\midrule
\endhead
\midrule\multicolumn{10}{r}{Continued on next page}\\
\endfoot
\bottomrule\endlastfoot
jewel2 q5 & 40 & \textbf{90.0} & \shortstack[r]{\underline{87.5}\\{\scriptsize \textcolor{losscolor}{-2.5}}} & \shortstack[r]{\textbf{90.0}\\{\scriptsize \textcolor{neutralcolor}{0.0}}} & \underline{87.5} & \shortstack[r]{\textbf{90.0}\\{\scriptsize \textcolor{gaincolor}{+2.5}}} & \shortstack[r]{\underline{87.5}\\{\scriptsize \textcolor{neutralcolor}{0.0}}} & \shortstack[r]{\underline{87.5}\\{\scriptsize \textcolor{neutralcolor}{0.0}}} & \cellcolor{methodshade}\shortstack[r]{\underline{87.5}\\{\scriptsize \textcolor{neutralcolor}{0.0}}}\\
jewel2 q6 & 20 & 60.0 & \shortstack[r]{65.0\\{\scriptsize \textcolor{gaincolor}{+5.0}}} & \shortstack[r]{70.0\\{\scriptsize \textcolor{gaincolor}{+10.0}}} & \underline{75.0} & \shortstack[r]{\underline{75.0}\\{\scriptsize \textcolor{neutralcolor}{0.0}}} & \shortstack[r]{70.0\\{\scriptsize \textcolor{losscolor}{-5.0}}} & \shortstack[r]{65.0\\{\scriptsize \textcolor{losscolor}{-10.0}}} & \cellcolor{methodshade}\shortstack[r]{\textbf{80.0}\\{\scriptsize \textcolor{gaincolor}{+5.0}}}\\
maze q4 & 35 & 51.4 & \shortstack[r]{45.7\\{\scriptsize \textcolor{losscolor}{-5.7}}} & \shortstack[r]{60.0\\{\scriptsize \textcolor{gaincolor}{+8.6}}} & 45.7 & \shortstack[r]{\underline{77.1}\\{\scriptsize \textcolor{gaincolor}{+31.4}}} & \shortstack[r]{60.0\\{\scriptsize \textcolor{gaincolor}{+14.3}}} & \shortstack[r]{57.1\\{\scriptsize \textcolor{gaincolor}{+11.4}}} & \cellcolor{methodshade}\shortstack[r]{\textbf{100.0}\\{\scriptsize \textcolor{gaincolor}{+54.3}}}\\
\end{longtable}
\endgroup

\begingroup
\small\setlength{\tabcolsep}{3pt}
\begin{longtable}{@{}lrrrrrrrrr@{}}
\caption{\textbf{GameQA-17: Gemini 3.1 Flash-Lite.} Complete-family accuracy (\%).}\label{tab:family_gameqa_flashlite}\\
\toprule
Family & $n$ & Dir. & GEPA & TG & Tool & XS & T2S & SO & \cellcolor{methodshade}VC\\\midrule
\endfirsthead
\multicolumn{10}{l}{\small\tablename~\thetable{} (continued)}\\
\toprule
Family & $n$ & Dir. & GEPA & TG & Tool & XS & T2S & SO & \cellcolor{methodshade}VC\\\midrule
\endhead
\midrule\multicolumn{10}{r}{Continued on next page}\\
\endfoot
\bottomrule\endlastfoot
jewel2 q5 & 40 & \textbf{72.5} & \shortstack[r]{62.5\\{\scriptsize \textcolor{losscolor}{-10.0}}} & \shortstack[r]{32.5\\{\scriptsize \textcolor{losscolor}{-40.0}}} & 30.0 & \shortstack[r]{57.5\\{\scriptsize \textcolor{gaincolor}{+27.5}}} & \shortstack[r]{52.5\\{\scriptsize \textcolor{gaincolor}{+22.5}}} & \shortstack[r]{\underline{70.0}\\{\scriptsize \textcolor{gaincolor}{+40.0}}} & \cellcolor{methodshade}\shortstack[r]{65.0\\{\scriptsize \textcolor{gaincolor}{+35.0}}}\\
jewel2 q6 & 20 & 20.0 & \shortstack[r]{35.0\\{\scriptsize \textcolor{gaincolor}{+15.0}}} & \shortstack[r]{30.0\\{\scriptsize \textcolor{gaincolor}{+10.0}}} & \underline{45.0} & \shortstack[r]{\textbf{50.0}\\{\scriptsize \textcolor{gaincolor}{+5.0}}} & \shortstack[r]{\underline{45.0}\\{\scriptsize \textcolor{neutralcolor}{0.0}}} & \shortstack[r]{40.0\\{\scriptsize \textcolor{losscolor}{-5.0}}} & \cellcolor{methodshade}\shortstack[r]{\textbf{50.0}\\{\scriptsize \textcolor{gaincolor}{+5.0}}}\\
maze q4 & 35 & 20.0 & \shortstack[r]{2.9\\{\scriptsize \textcolor{losscolor}{-17.1}}} & \shortstack[r]{25.7\\{\scriptsize \textcolor{gaincolor}{+5.7}}} & 14.3 & \shortstack[r]{14.3\\{\scriptsize \textcolor{neutralcolor}{0.0}}} & \shortstack[r]{8.6\\{\scriptsize \textcolor{losscolor}{-5.7}}} & \shortstack[r]{\underline{28.6}\\{\scriptsize \textcolor{gaincolor}{+14.3}}} & \cellcolor{methodshade}\shortstack[r]{\textbf{100.0}\\{\scriptsize \textcolor{gaincolor}{+85.7}}}\\
minesweeper q5 & 40 & 50.0 & \shortstack[r]{\underline{82.5}\\{\scriptsize \textcolor{gaincolor}{+32.5}}} & \shortstack[r]{75.0\\{\scriptsize \textcolor{gaincolor}{+25.0}}} & 55.0 & \shortstack[r]{70.0\\{\scriptsize \textcolor{gaincolor}{+15.0}}} & \shortstack[r]{62.5\\{\scriptsize \textcolor{gaincolor}{+7.5}}} & \shortstack[r]{72.5\\{\scriptsize \textcolor{gaincolor}{+17.5}}} & \cellcolor{methodshade}\shortstack[r]{\textbf{85.0}\\{\scriptsize \textcolor{gaincolor}{+30.0}}}\\
\end{longtable}
\endgroup

\begingroup
\small\setlength{\tabcolsep}{3pt}
\begin{longtable}{@{}lrrrrr@{}}
\caption{\textbf{GameQA-17: DeepSeek and GPT with frozen FL routes.} Complete-family accuracy (\%).}\label{tab:family_gameqa_transfer}\\
\toprule
Family & $n$ & DS Tool & \cellcolor{methodshade}DS VC & GPT Tool & \cellcolor{methodshade}GPT VC\\\midrule
\endfirsthead
\multicolumn{6}{l}{\small\tablename~\thetable{} (continued)}\\
\toprule
Family & $n$ & DS Tool & \cellcolor{methodshade}DS VC & GPT Tool & \cellcolor{methodshade}GPT VC\\\midrule
\endhead
\midrule\multicolumn{6}{r}{Continued on next page}\\
\endfoot
\bottomrule\endlastfoot
jewel2 q5 & 40 & \underline{55.0} & \cellcolor{methodshade}\shortstack[r]{\textbf{65.0}\\{\scriptsize \textcolor{gaincolor}{+10.0}}} & \underline{60.0} & \cellcolor{methodshade}\shortstack[r]{\textbf{65.0}\\{\scriptsize \textcolor{gaincolor}{+5.0}}}\\
jewel2 q6 & 20 & \textbf{60.0} & \cellcolor{methodshade}\shortstack[r]{\textbf{60.0}\\{\scriptsize \textcolor{neutralcolor}{0.0}}} & \textbf{35.0} & \cellcolor{methodshade}\shortstack[r]{\underline{20.0}\\{\scriptsize \textcolor{losscolor}{-15.0}}}\\
maze q4 & 35 & \underline{97.1} & \cellcolor{methodshade}\shortstack[r]{\textbf{100.0}\\{\scriptsize \textcolor{gaincolor}{+2.9}}} & \underline{34.3} & \cellcolor{methodshade}\shortstack[r]{\textbf{100.0}\\{\scriptsize \textcolor{gaincolor}{+65.7}}}\\
minesweeper q5 & 40 & \underline{70.0} & \cellcolor{methodshade}\shortstack[r]{\textbf{85.0}\\{\scriptsize \textcolor{gaincolor}{+15.0}}} & \underline{35.0} & \cellcolor{methodshade}\shortstack[r]{\textbf{72.5}\\{\scriptsize \textcolor{gaincolor}{+37.5}}}\\
pacman q1 & 20 & \textbf{95.0} & \cellcolor{methodshade}\shortstack[r]{\textbf{95.0}\\{\scriptsize \textcolor{neutralcolor}{0.0}}} & \underline{5.0} & \cellcolor{methodshade}\shortstack[r]{\textbf{95.0}\\{\scriptsize \textcolor{gaincolor}{+90.0}}}\\
pacman q3 & 20 & \underline{70.0} & \cellcolor{methodshade}\shortstack[r]{\textbf{100.0}\\{\scriptsize \textcolor{gaincolor}{+30.0}}} & \underline{0.0} & \cellcolor{methodshade}\shortstack[r]{\textbf{100.0}\\{\scriptsize \textcolor{gaincolor}{+100.0}}}\\
pacman q4 & 20 & \underline{70.0} & \cellcolor{methodshade}\shortstack[r]{\textbf{75.0}\\{\scriptsize \textcolor{gaincolor}{+5.0}}} & \underline{10.0} & \cellcolor{methodshade}\shortstack[r]{\textbf{15.0}\\{\scriptsize \textcolor{gaincolor}{+5.0}}}\\
\end{longtable}
\endgroup

\begingroup
\small\setlength{\tabcolsep}{3pt}
\begin{longtable}{@{}lrrrrrrrrr@{}}
\caption{\textbf{PuzzleVQA: Gemini 3.1 Pro.} Complete-family accuracy (\%).}\label{tab:family_puzzlevqa_pro}\\
\toprule
Family & $n$ & Dir. & GEPA & TG & Tool & XS & T2S & SO & \cellcolor{methodshade}VC\\\midrule
\endfirsthead
\multicolumn{10}{l}{\small\tablename~\thetable{} (continued)}\\
\toprule
Family & $n$ & Dir. & GEPA & TG & Tool & XS & T2S & SO & \cellcolor{methodshade}VC\\\midrule
\endhead
\midrule\multicolumn{10}{r}{Continued on next page}\\
\endfoot
\bottomrule\endlastfoot
circle size number & 67 & 70.1 & \shortstack[r]{\underline{95.5}\\{\scriptsize \textcolor{gaincolor}{+25.4}}} & \shortstack[r]{\textbf{100.0}\\{\scriptsize \textcolor{gaincolor}{+29.9}}} & 77.6 & \shortstack[r]{\textbf{100.0}\\{\scriptsize \textcolor{gaincolor}{+22.4}}} & \shortstack[r]{88.1\\{\scriptsize \textcolor{gaincolor}{+10.4}}} & \shortstack[r]{\textbf{100.0}\\{\scriptsize \textcolor{gaincolor}{+22.4}}} & \cellcolor{methodshade}\shortstack[r]{\textbf{100.0}\\{\scriptsize \textcolor{gaincolor}{+22.4}}}\\
color grid & 67 & \textbf{100.0} & \shortstack[r]{\textbf{100.0}\\{\scriptsize \textcolor{neutralcolor}{0.0}}} & \shortstack[r]{\textbf{100.0}\\{\scriptsize \textcolor{neutralcolor}{0.0}}} & \textbf{100.0} & \shortstack[r]{\textbf{100.0}\\{\scriptsize \textcolor{neutralcolor}{0.0}}} & \shortstack[r]{\textbf{100.0}\\{\scriptsize \textcolor{neutralcolor}{0.0}}} & \shortstack[r]{\textbf{100.0}\\{\scriptsize \textcolor{neutralcolor}{0.0}}} & \cellcolor{methodshade}\shortstack[r]{\textbf{100.0}\\{\scriptsize \textcolor{neutralcolor}{0.0}}}\\
color hexagon & 67 & \textbf{100.0} & \shortstack[r]{\textbf{100.0}\\{\scriptsize \textcolor{neutralcolor}{0.0}}} & \shortstack[r]{\textbf{100.0}\\{\scriptsize \textcolor{neutralcolor}{0.0}}} & \textbf{100.0} & \shortstack[r]{\textbf{100.0}\\{\scriptsize \textcolor{neutralcolor}{0.0}}} & \shortstack[r]{\textbf{100.0}\\{\scriptsize \textcolor{neutralcolor}{0.0}}} & \shortstack[r]{\textbf{100.0}\\{\scriptsize \textcolor{neutralcolor}{0.0}}} & \cellcolor{methodshade}\shortstack[r]{\textbf{100.0}\\{\scriptsize \textcolor{neutralcolor}{0.0}}}\\
\end{longtable}
\endgroup

\begingroup
\small\setlength{\tabcolsep}{3pt}
\begin{longtable}{@{}lrrrrrrrrr@{}}
\caption{\textbf{PuzzleVQA: Gemini 3.1 Flash-Lite.} Complete-family accuracy (\%).}\label{tab:family_puzzlevqa_flashlite}\\
\toprule
Family & $n$ & Dir. & GEPA & TG & Tool & XS & T2S & SO & \cellcolor{methodshade}VC\\\midrule
\endfirsthead
\multicolumn{10}{l}{\small\tablename~\thetable{} (continued)}\\
\toprule
Family & $n$ & Dir. & GEPA & TG & Tool & XS & T2S & SO & \cellcolor{methodshade}VC\\\midrule
\endhead
\midrule\multicolumn{10}{r}{Continued on next page}\\
\endfoot
\bottomrule\endlastfoot
circle size number & 67 & 74.6 & \shortstack[r]{\textbf{100.0}\\{\scriptsize \textcolor{gaincolor}{+25.4}}} & \shortstack[r]{\textbf{100.0}\\{\scriptsize \textcolor{gaincolor}{+25.4}}} & 71.6 & \shortstack[r]{\underline{92.5}\\{\scriptsize \textcolor{gaincolor}{+20.9}}} & \shortstack[r]{71.6\\{\scriptsize \textcolor{neutralcolor}{0.0}}} & \shortstack[r]{\textbf{100.0}\\{\scriptsize \textcolor{gaincolor}{+28.4}}} & \cellcolor{methodshade}\shortstack[r]{\textbf{100.0}\\{\scriptsize \textcolor{gaincolor}{+28.4}}}\\
color grid & 67 & 79.1 & \shortstack[r]{92.5\\{\scriptsize \textcolor{gaincolor}{+13.4}}} & \shortstack[r]{79.1\\{\scriptsize \textcolor{neutralcolor}{0.0}}} & 95.5 & \shortstack[r]{\textbf{100.0}\\{\scriptsize \textcolor{gaincolor}{+4.5}}} & \shortstack[r]{\underline{97.0}\\{\scriptsize \textcolor{gaincolor}{+1.5}}} & \shortstack[r]{\textbf{100.0}\\{\scriptsize \textcolor{gaincolor}{+4.5}}} & \cellcolor{methodshade}\shortstack[r]{\textbf{100.0}\\{\scriptsize \textcolor{gaincolor}{+4.5}}}\\
color hexagon & 67 & \textbf{100.0} & \shortstack[r]{\textbf{100.0}\\{\scriptsize \textcolor{neutralcolor}{0.0}}} & \shortstack[r]{\textbf{100.0}\\{\scriptsize \textcolor{neutralcolor}{0.0}}} & \textbf{100.0} & \shortstack[r]{\textbf{100.0}\\{\scriptsize \textcolor{neutralcolor}{0.0}}} & \shortstack[r]{\textbf{100.0}\\{\scriptsize \textcolor{neutralcolor}{0.0}}} & \shortstack[r]{\textbf{100.0}\\{\scriptsize \textcolor{neutralcolor}{0.0}}} & \cellcolor{methodshade}\shortstack[r]{\textbf{100.0}\\{\scriptsize \textcolor{neutralcolor}{0.0}}}\\
\end{longtable}
\endgroup

\begingroup
\small\setlength{\tabcolsep}{3pt}
\begin{longtable}{@{}lrrrrr@{}}
\caption{\textbf{PuzzleVQA: DeepSeek and GPT with frozen FL routes.} Complete-family accuracy (\%).}\label{tab:family_puzzlevqa_transfer}\\
\toprule
Family & $n$ & DS Tool & \cellcolor{methodshade}DS VC & GPT Tool & \cellcolor{methodshade}GPT VC\\\midrule
\endfirsthead
\multicolumn{6}{l}{\small\tablename~\thetable{} (continued)}\\
\toprule
Family & $n$ & DS Tool & \cellcolor{methodshade}DS VC & GPT Tool & \cellcolor{methodshade}GPT VC\\\midrule
\endhead
\midrule\multicolumn{6}{r}{Continued on next page}\\
\endfoot
\bottomrule\endlastfoot
circle size number & 67 & \underline{74.6} & \cellcolor{methodshade}\shortstack[r]{\textbf{100.0}\\{\scriptsize \textcolor{gaincolor}{+25.4}}} & \underline{29.9} & \cellcolor{methodshade}\shortstack[r]{\textbf{100.0}\\{\scriptsize \textcolor{gaincolor}{+70.1}}}\\
color grid & 67 & \underline{95.5} & \cellcolor{methodshade}\shortstack[r]{\textbf{100.0}\\{\scriptsize \textcolor{gaincolor}{+4.5}}} & \underline{52.2} & \cellcolor{methodshade}\shortstack[r]{\textbf{100.0}\\{\scriptsize \textcolor{gaincolor}{+47.8}}}\\
color hexagon & 67 & \textbf{100.0} & \cellcolor{methodshade}\shortstack[r]{\underline{98.5}\\{\scriptsize \textcolor{losscolor}{-1.5}}} & \underline{17.9} & \cellcolor{methodshade}\shortstack[r]{\textbf{38.8}\\{\scriptsize \textcolor{gaincolor}{+20.9}}}\\
color number hexagon & 67 & \underline{91.0} & \cellcolor{methodshade}\shortstack[r]{\textbf{98.5}\\{\scriptsize \textcolor{gaincolor}{+7.5}}} & \underline{37.3} & \cellcolor{methodshade}\shortstack[r]{\textbf{97.0}\\{\scriptsize \textcolor{gaincolor}{+59.7}}}\\
color overlap squares & 67 & \underline{82.1} & \cellcolor{methodshade}\shortstack[r]{\textbf{88.1}\\{\scriptsize \textcolor{gaincolor}{+6.0}}} & \underline{14.9} & \cellcolor{methodshade}\shortstack[r]{\textbf{37.3}\\{\scriptsize \textcolor{gaincolor}{+22.4}}}\\
color size circle & 66 & \underline{90.9} & \cellcolor{methodshade}\shortstack[r]{\textbf{100.0}\\{\scriptsize \textcolor{gaincolor}{+9.1}}} & \underline{47.0} & \cellcolor{methodshade}\shortstack[r]{\textbf{100.0}\\{\scriptsize \textcolor{gaincolor}{+53.0}}}\\
grid number & 67 & \underline{95.5} & \cellcolor{methodshade}\shortstack[r]{\textbf{98.5}\\{\scriptsize \textcolor{gaincolor}{+3.0}}} & \underline{62.7} & \cellcolor{methodshade}\shortstack[r]{\textbf{98.5}\\{\scriptsize \textcolor{gaincolor}{+35.8}}}\\
\end{longtable}
\endgroup

\subsection{Complete-Family Ablation Scores}

\label{app:family_ablations}

The following tables expand Table~\ref{tab:ablations} over all 12 TIR and 20 MIRA families, including unaffected families. $\ast$ marks the five re-acquired families. Other families retain the frozen full policy. The acquisition and deployment comparisons use different artifacts. Deployment TEXT/CODE/INJECT replace only the affected program takeovers of the full policy; all other outputs stay fixed. SCALAR supplements the main acquisition comparison with aggregate feedback. The no-text-fallback condition removes the trajectory-derived text fallback. Both benchmarks use the same fixed single-rollout Tool and fallback records as the main comparison. All acquisition entries report mean accuracy over three independent acquisition runs for every re-acquired family. Dataset totals weight these family scores by the number of questions.

\begingroup
\small\setlength{\tabcolsep}{3pt}
\begin{longtable}{@{}lrrrrrr@{}}
\caption{\textbf{TIR-Bench: complete-family acquisition ablations.} Accuracy (\%).}\label{tab:family_tirbench_acquisition}\\
\toprule
Family & $n$ & Tool & FULLC & IID & SCALAR & No scope\\\midrule
\endfirsthead
\multicolumn{7}{l}{\small\tablename~\thetable{} (continued)}\\
\toprule
Family & $n$ & Tool & FULLC & IID & SCALAR & No scope\\\midrule
\endhead
\midrule\multicolumn{7}{r}{Continued on next page}\\
\endfoot
\bottomrule\endlastfoot
color & 67 & 74.6 & 70.1 & 70.1 & 70.1 & 70.1\\
contrast & 34 & 64.7 & 52.9 & 52.9 & 52.9 & 52.9\\
instrument & 54 & 24.1 & 25.9 & 25.9 & 25.9 & 25.9\\
jigsaw$^\ast$ & 80 & 5.0 & 90.4 & 75.0 & 47.1 & 81.2\\
math & 80 & 77.5 & 81.2 & 81.2 & 81.2 & 81.2\\
maze$^\ast$ & 80 & 90.0 & 96.7 & 97.1 & 90.4 & 43.8\\
ocr & 40 & 55.0 & 87.5 & 87.5 & 87.5 & 87.5\\
refcoco$^\ast$ & 80 & 47.5 & 52.1 & 47.5 & 47.5 & 54.2\\
rotation game & 50 & 92.0 & 100.0 & 100.0 & 100.0 & 100.0\\
symbolic$^\ast$ & 34 & 88.2 & 92.2 & 88.2 & 88.2 & 85.3\\
visual search & 80 & 70.0 & 70.0 & 70.0 & 70.0 & 70.0\\
word search$^\ast$ & 67 & 80.6 & 87.6 & 86.1 & 85.1 & 78.1\\
\midrule
Dataset total & 746 & 62.9 & 75.9 & 73.5 & 69.7 & 68.3\\
\end{longtable}
\endgroup

\begingroup
\small\setlength{\tabcolsep}{3pt}
\begin{longtable}{@{}lrrrrrrr@{}}
\caption{\textbf{TIR-Bench: complete-family deployment ablations.} Accuracy (\%).}\label{tab:family_tirbench_deployment}\\
\toprule
Family & $n$ & Tool & Full & TEXT & CODE & INJECT & \shortstack[r]{No text\\fallback}\\\midrule
\endfirsthead
\multicolumn{8}{l}{\small\tablename~\thetable{} (continued)}\\
\toprule
Family & $n$ & Tool & Full & TEXT & CODE & INJECT & \shortstack[r]{No text\\fallback}\\\midrule
\endhead
\midrule\multicolumn{8}{r}{Continued on next page}\\
\endfoot
\bottomrule\endlastfoot
color & 67 & 74.6 & 70.1 & 70.1 & 70.1 & 70.1 & 74.6\\
contrast & 34 & 64.7 & 52.9 & 52.9 & 52.9 & 52.9 & 64.7\\
instrument & 54 & 24.1 & 25.9 & 25.9 & 25.9 & 25.9 & 24.1\\
jigsaw & 80 & 5.0 & 95.0 & 33.8 & 72.5 & 92.5 & 95.0\\
math & 80 & 77.5 & 81.2 & 81.2 & 81.2 & 81.2 & 77.5\\
maze & 80 & 90.0 & 100.0 & 97.5 & 100.0 & 100.0 & 100.0\\
ocr & 40 & 55.0 & 87.5 & 87.5 & 87.5 & 82.5 & 87.5\\
refcoco & 80 & 47.5 & 55.0 & 55.0 & 55.0 & 55.0 & 47.5\\
rotation game & 50 & 92.0 & 100.0 & 100.0 & 90.0 & 100.0 & 100.0\\
symbolic & 34 & 88.2 & 91.2 & 91.2 & 91.2 & 85.3 & 91.2\\
visual search & 80 & 70.0 & 70.0 & 70.0 & 70.0 & 70.0 & 70.0\\
word search & 67 & 80.6 & 79.1 & 79.1 & 79.1 & 79.1 & 80.6\\
\midrule
Dataset total & 746 & 62.9 & 76.3 & 69.4 & 73.2 & 75.5 & 76.0\\
\end{longtable}
\endgroup

\begingroup
\small\setlength{\tabcolsep}{3pt}
\begin{longtable}{@{}lrrrrrr@{}}
\caption{\textbf{MIRA: complete-family acquisition ablations.} Accuracy (\%).}\label{tab:family_mira_acquisition}\\
\toprule
Family & $n$ & Tool & FULLC & IID & SCALAR & No scope\\\midrule
\endfirsthead
\multicolumn{7}{l}{\small\tablename~\thetable{} (continued)}\\
\toprule
Family & $n$ & Tool & FULLC & IID & SCALAR & No scope\\\midrule
\endhead
\midrule\multicolumn{7}{r}{Continued on next page}\\
\endfoot
\bottomrule\endlastfoot
billiards$^\ast$ & 14 & 35.7 & 47.6 & 42.9 & 61.9 & 50.0\\
convex hull & 20 & 90.0 & 90.0 & 90.0 & 90.0 & 90.0\\
cubes count & 20 & 5.0 & 5.0 & 5.0 & 5.0 & 5.0\\
cubes missing & 20 & 5.0 & 0.0 & 0.0 & 0.0 & 0.0\\
defuse a bomb & 17 & 23.5 & 17.6 & 17.6 & 17.6 & 17.6\\
electric charge & 14 & 78.6 & 85.7 & 85.7 & 85.7 & 85.7\\
gear rotation & 26 & 96.2 & 100.0 & 100.0 & 100.0 & 100.0\\
localizer & 16 & 0.0 & 0.0 & 0.0 & 0.0 & 0.0\\
mirror clock$^\ast$ & 20 & 75.0 & 88.3 & 83.3 & 80.0 & 90.0\\
mirror pattern$^\ast$ & 20 & 20.0 & 98.3 & 96.7 & 95.0 & 25.0\\
multi piece puzzle & 20 & 5.0 & 5.0 & 5.0 & 5.0 & 5.0\\
overlap & 20 & 65.0 & 60.0 & 60.0 & 60.0 & 60.0\\
paper airplane & 17 & 17.6 & 29.4 & 29.4 & 29.4 & 29.4\\
puzzle$^\ast$ & 17 & 52.9 & 52.9 & 52.9 & 52.9 & 13.7\\
rolling dice sum & 17 & 5.9 & 58.8 & 58.8 & 58.8 & 58.8\\
rolling dice top & 17 & 29.4 & 47.1 & 47.1 & 47.1 & 47.1\\
rolling dice two & 17 & 0.0 & 5.9 & 5.9 & 5.9 & 5.9\\
trailer cubes count$^\ast$ & 17 & 70.6 & 70.6 & 70.6 & 70.6 & 94.1\\
trailer cubes missing & 17 & 0.0 & 0.0 & 0.0 & 0.0 & 0.0\\
unfolded cube & 17 & 41.2 & 41.2 & 41.2 & 41.2 & 41.2\\
\midrule
Dataset total & 363 & 37.2 & 46.6 & 46.0 & 46.5 & 42.0\\
\end{longtable}
\endgroup

\begingroup
\small\setlength{\tabcolsep}{3pt}
\begin{longtable}{@{}lrrrrrrr@{}}
\caption{\textbf{MIRA: complete-family deployment ablations.} Accuracy (\%).}\label{tab:family_mira_deployment}\\
\toprule
Family & $n$ & Tool & Full & TEXT & CODE & INJECT & \shortstack[r]{No text\\fallback}\\\midrule
\endfirsthead
\multicolumn{8}{l}{\small\tablename~\thetable{} (continued)}\\
\toprule
Family & $n$ & Tool & Full & TEXT & CODE & INJECT & \shortstack[r]{No text\\fallback}\\\midrule
\endhead
\midrule\multicolumn{8}{r}{Continued on next page}\\
\endfoot
\bottomrule\endlastfoot
billiards & 14 & 35.7 & 71.4 & 57.1 & 57.1 & 71.4 & 71.4\\
convex hull & 20 & 90.0 & 90.0 & 90.0 & 90.0 & 90.0 & 90.0\\
cubes count & 20 & 5.0 & 5.0 & 5.0 & 5.0 & 5.0 & 5.0\\
cubes missing & 20 & 5.0 & 0.0 & 0.0 & 0.0 & 0.0 & 5.0\\
defuse a bomb & 17 & 23.5 & 17.6 & 17.6 & 17.6 & 17.6 & 23.5\\
electric charge & 14 & 78.6 & 85.7 & 85.7 & 85.7 & 85.7 & 78.6\\
gear rotation & 26 & 96.2 & 100.0 & 100.0 & 100.0 & 100.0 & 96.2\\
localizer & 16 & 0.0 & 0.0 & 0.0 & 0.0 & 0.0 & 0.0\\
mirror clock & 20 & 75.0 & 100.0 & 80.0 & 70.0 & 80.0 & 100.0\\
mirror pattern & 20 & 20.0 & 100.0 & 80.0 & 90.0 & 85.0 & 100.0\\
multi piece puzzle & 20 & 5.0 & 5.0 & 5.0 & 5.0 & 5.0 & 5.0\\
overlap & 20 & 65.0 & 60.0 & 60.0 & 60.0 & 60.0 & 65.0\\
paper airplane & 17 & 17.6 & 29.4 & 29.4 & 29.4 & 29.4 & 17.6\\
puzzle & 17 & 52.9 & 23.5 & 23.5 & 35.3 & 29.4 & 23.5\\
rolling dice sum & 17 & 5.9 & 58.8 & 58.8 & 58.8 & 58.8 & 58.8\\
rolling dice top & 17 & 29.4 & 47.1 & 47.1 & 47.1 & 47.1 & 47.1\\
rolling dice two & 17 & 0.0 & 5.9 & 5.9 & 5.9 & 5.9 & 0.0\\
trailer cubes count & 17 & 70.6 & 76.5 & 76.5 & 76.5 & 76.5 & 70.6\\
trailer cubes missing & 17 & 0.0 & 0.0 & 0.0 & 0.0 & 0.0 & 0.0\\
unfolded cube & 17 & 41.2 & 41.2 & 41.2 & 41.2 & 41.2 & 41.2\\
\midrule
Dataset total & 363 & 37.2 & 47.1 & 44.4 & 44.9 & 45.5 & 46.3\\
\end{longtable}
\endgroup

\paragraph{Related protocols and sampling results.} Acquisition and deployment protocols appear in Appendices~\ref{app:tir_results} and~\ref{app:mira_results}; Table~\ref{tab:tir_sampling_families} reports all twelve TIR family sampling scores under their separate four-attempt sampling protocol.